\documentclass[10pt,journal,compsoc]{IEEEtran}
\usepackage{cite}
\usepackage{amsmath,amssymb,amsfonts}
\usepackage{algorithmic}
\usepackage{graphicx}
\usepackage{textcomp}
\usepackage[table]{ xcolor}
\usepackage{xcolor}

\usepackage[utf8]{inputenc} % allow utf-8 input
\usepackage[T1]{fontenc}    % use 8-bit T1 fonts
\usepackage{hyperref}       % hyperlinks
\usepackage{url}            % simple URL typesetting
\usepackage{booktabs}       % professional-quality tables
\usepackage{amsfonts}       % blackboard math symbols
\usepackage{nicefrac}       % compact symbols for 1/2, etc.
\usepackage{microtype}      % microtypography

\usepackage{bm}
\usepackage{comment}
\usepackage{amsmath}
\usepackage{enumerate}%\begin{enumerate}[i] % reacts to 1, a, A, i, and I
\usepackage{enumitem}%\begin{enumerate}[i] % reacts to 1, a, A, i, and I
\usepackage{mathrsfs}
\usepackage{makeidx}         % allows index generation
\usepackage{graphicx}        % standard LaTeX graphics tool
\usepackage{multicol}        % used for the two-column index
\usepackage{subfigure}
\usepackage{epsfig,amssymb,latexsym}
\usepackage{psfrag}
\usepackage{algorithmic}
\usepackage{algorithm}
\usepackage{makecell}

\usepackage{colortbl}
\definecolor{lightgray}{rgb}{.92,.92,.92}
\definecolor{lavender}{rgb}{0.70,0.70,1.00}
\definecolor{deepred}{rgb}{1.0,0.0,0.0}

\usepackage{fancyhdr}
\usepackage{multirow}

\usepackage{caption}

\usepackage{color}

\usepackage{environ}
\usepackage{tikz}
\NewEnviron{elaboration}{
\par
\begin{tikzpicture}
\node[rectangle,minimum width=0.370\textwidth] (m) {\begin{minipage}{0.38\textwidth}\BODY\end{minipage}};
\draw[dashed] (m.south west) rectangle (m.north east);
\end{tikzpicture}
}

\newcommand{\mr}{\mathrm}
\newcommand{\mc}{\mathcal}

\ifCLASSINFOpdf
\else
\fi

\begin{document}
\title{Create Your World: Lifelong Text-to-Image Diffusion}
%Qianqian~Wang,
\author{Gan~Sun,~\IEEEmembership{Member,~IEEE,}
        Wenqi~Liang\textsuperscript{†},
        Jiahua~Dong,
        Jun~Li,
        Zhengming Ding, 
        Yang~Cong\textsuperscript{*},~\IEEEmembership{Senior Member,~IEEE}
%        and~Yun~Fu,~\IEEEmembership{Fellow,~IEEE}% <-this % stops a space
\IEEEcompsocitemizethanks{\IEEEcompsocthanksitem Gan Sun, Wenqi Liang and Jiahua Dong are with State Key Laboratory of Robotics, Shenyang Institute of Automation, Institutes for Robotics and Intelligent Manufacturing, Chinese Academy of Sciences, Shenyang, 110016, China (email: sungan1412@gmail.com, liangwenqi0123@gmail.com, dongjiahua1995@gmail.com) \protect\\
\vspace{-8pt}
\IEEEcompsocthanksitem Jun Li is with the School of Computer Science and Engineering, Nanjing University of Science and Technology, Jiangsu 210094, China (e-mail: junli@njust.edu.cn). \protect\\
\IEEEcompsocthanksitem Zhengming Ding is with the Department of Computer Science, Tulane University, New Orleans, LA 70118, USA (e-mail: zding1@tulane.edu). \protect\\
\IEEEcompsocthanksitem Yang Cong is with the College of Automation Science and Engineering, South China University of Technology, Guangzhou 510640, China (email: congyang81@gmail.com)). \protect\\
\vspace{-8pt}
\IEEEcompsocthanksitem Wenqi Liang and Jiahua Dong are also with University of Chinese Academy of Sciences, Beijing, 100049, China. \protect\\
% and with the Department of Electrical and Computer Engineering, Northeastern University, Boston, MA. 02115 USA \protect\\
%\IEEEcompsocthanksitem Q. Wang is with Xidian University. Xian, Shanxi, 710071, China, Email: qianqian174@foxmail.com \protect\\
%\IEEEcompsocthanksitem B. Zhong is with Huaqiao University, Xiamen, Fujian, 361021, China, Email:  bnzhong@hqu.edu.cn \protect\\
%\IEEEcompsocthanksitem Y. Fu is with the Department of Electrical and Computer Engineering and the Khoury College of Computer and Information Sciences, Northeastern University, Boston, MA. 02115 USA. Email: yunfu@ece.neu.edu \protect
}% <-this % stops an unwanted space
\thanks{This work is supported by National Nature Science Foundation of China under Grant (62225310,62003336,62273333), CAS-Youth Innovation Promotion Association Scholarship under Grant 2023207.}
\thanks{†Equal contribution. *The corresponding author is Prof. Yang Cong.}
%\thanks{(Corresponding author: Prof. Yang Cong.)}
}

%Nature Foundation of Liaoning Province of China under Grant (2020-KF-11-01)

\markboth{Journal of \LaTeX\ Class Files,~Vol.~14, No.~8, August~2015}%
{Shell \MakeLowercase{\textit{et al.}}: Bare Demo of IEEEtran.cls for Computer Society Journals}


%selectively transferring accumulated nonlinear knowledge
\IEEEtitleabstractindextext{%
 \begin{center}\setcounter{figure}{0}
    \includegraphics[width =480pt]{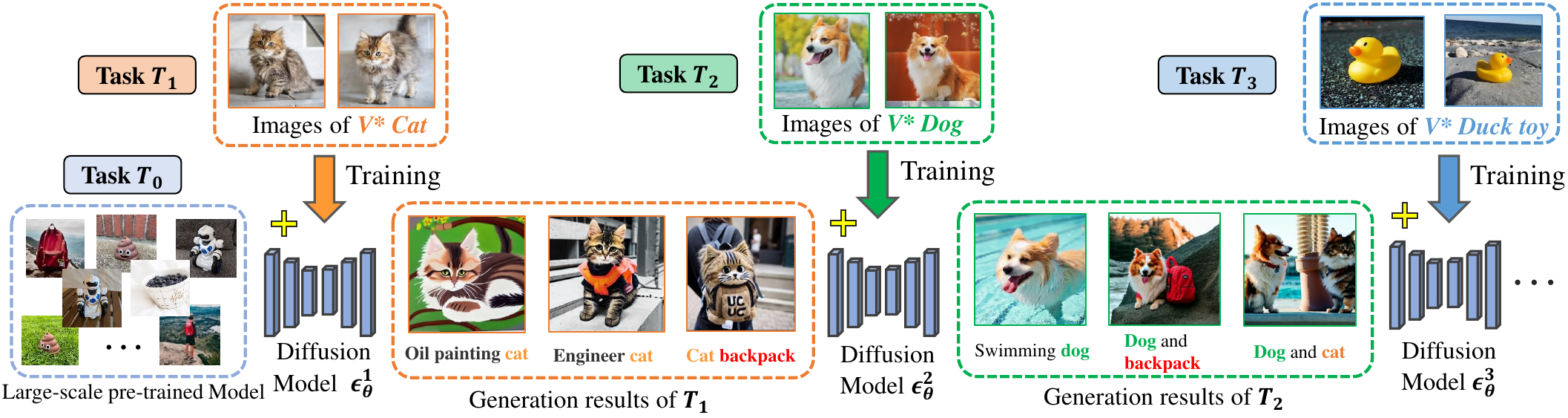}
     \captionof{figure}{Motivation illustration of our proposed lifelong text-to-image diffusion problem, where the large-scale pre-trained diffusion model should be gradually extended into a sequence of user-specific generation tasks, \emph{i.e.,} \textit{V*Cat, V*Dog and V*Duck toy}.} 
    \label{fig:motivation}
    \vspace{-5pt}
  \end{center}

\begin{abstract}
 Text-to-image generative models can produce diverse high-quality images of concepts with a text prompt, which have demonstrated excellent ability in image generation, image translation, etc. We in this work study the problem of synthesizing instantiations of a user's own concepts in a never-ending manner, \emph{i.e.,} {create your world}, where the new concepts from user are quickly learned with a few examples. To achieve this goal, we propose a  \underline{L}ife\underline{l}ong {t}ext-to-{i}mage \underline{D}iffusion \underline{M}odel (L$^2$DM), which intends to overcome knowledge ``catastrophic forgetting'' for the past encountered concepts, and semantic ``catastrophic neglecting'' for one or more concepts in the text prompt. In respect of knowledge ``catastrophic forgetting'', our L$^2$DM framework devises a task-aware memory enhancement module and a elastic-concept distillation module, which could respectively safeguard the knowledge of both prior concepts and each past personalized concept. When generating images with a user text prompt, the solution to semantic ``catastrophic neglecting'' is that a concept attention artist module can alleviate the semantic neglecting from concept aspect, and an orthogonal attention module can reduce the semantic binding from attribute aspect. To the end, our model can generate more faithful image across a range of continual text prompts in terms of both qualitative and quantitative metrics, when comparing with the related state-of-the-art models. 
 The code will be released at \url{https://wenqiliang.github.io/}. 

%the performance of the proposed framework in terms of effectiveness and efficiency, which validates
\end{abstract}

\begin{IEEEkeywords}
Lifelong Machine Learning, Stable Diffusion, Image Generation, Continual Learning.
\end{IEEEkeywords}}

\maketitle

\IEEEdisplaynontitleabstractindextext

\IEEEpeerreviewmaketitle

%to discover the corresponding embedding of data, which has shown the state-of-the-art performance in many applications

\IEEEraisesectionheading{\section{Introduction}\label{sec:introduction}}
\IEEEPARstart{R}ecently-proposed text-to-image diffusion models~\cite{xu2023open,blattmann2023align,gu2022vector,balaji2022ediffi} have shown the impressive development amongst computer vision applications, which have raised the performance of general generative models to a new level. With the unparalleled generation ability, the users can generate creative and high-quality image via a free-form text prompt. For instance, the classical Stable Diffusion method \cite{rombach2022high} turns diffusion models into powerful and flexible generators for general conditioning inputs, which achieves the competing results for image inpainting, text-to-image synthesis, super-resolution and so on; \cite{wang2022diffusiondb} introduces the first large-scale text-to-image prompt dataset, which can analyze prompts and discuss key properties or patterns of key prompts.

\begin{comment}
\begin{figure*}[t]
\centering
\includegraphics[width =480pt ,height =124pt]{fig//motivation_new.pdf}
\vspace{-4pt}
\caption{Illustration of our generalized lifelong spectral clustering (GL$^2$SC) model, where different shapes in matrices $X^t$ and $W^t$ are from different clusters. When observing a new clustering task $t$, the knowledge or experience is iteratively transferred from orthogonal basis memory and feature embedding memory to encode the new task.}
\vspace{-12pt}
\label{fig:lifelongspectralclustering}
\end{figure*}
\end{comment}

% introduce the motivation for the create my world

However, most recent models~\cite{nichol2021glide,saharia2022photorealistic} focus on how to boost the generation performance with preset large-scale dataset and fixed concepts. These models cannot be generalized into unseen concepts provided by the user, or personalizing the generation model. As shown in Fig.~\ref{fig:motivation}, when users wish to synthesize specific concepts by using their own personal lives in a never-ending manner, these above models could cause high computational consumption and model overfit in real-world applications. For example, after incorporating a large-scale diffusion model with personal pets such as loved \textit{dog}, \emph{e.g.,} ``\textit{a photo of pet dog}", the user can compose this loved dog with existing prior concepts to synthesize new variations. When the user tries to augment this diffusion model with his/her friend \textit{James} in the next day, ``\textit{a photo of pet dog with James}'', a straightforward manner is fine-tuning a large-scale diffusion model with both loved \textit{dog} and \textit{James}. However, this manner could present a realistic computational cost and further render the fine-tuning process impracticable, when the user needs to learn new concepts with the pre-trained large-scale diffusion model in a never-ending way.

%  introduce the challenges  for lifelong text-to-image generation and two problems
Motivated by the aforementioned scenario, here this paper explores how to establish a generalized lifelong text-to-image diffusion model, which aims to consecutively compose multiple personalized concepts of user while maintaining the original distinctive features. To achieve this, we consider the following challenges associated with lifelong text-to-image diffusion problems:
\vspace{-1pt}
\begin{itemize}[leftmargin=12pt]
\setlength{\itemsep}{2pt}
\setlength{\parsep}{0pt}
\setlength{\parskip}{0pt}
      %the hierarchical cluster space among past and coming clustering tasks should be consistent.
 
\item ``\textbf{Catastrophic Forgetting}'', \emph{i.e.,} the diffusion model tends to catastrophic forget the knowledge or meaning of encountered concepts. For this respect, two types of knowledge forgetting should be concerned: 1) prior knowledge forgetting, \emph{i.e.,} the generalization ability of the large-scale diffusion model could be inconsistent with the original one after learning personalized concepts. For example, the meaning of ``moon'' could be inconsistent when adding the new concept ``moongate''. 2) personalized knowledge forgetting, \emph{i.e.,} the ability to backtrack the specific concepts provided by users. For instance, the user could compose the earlier concept loved ``dog'' without using training data, even after a sequence of specific concepts.

      %the cluster centers of the news clustering task for year 2010 could be \{Entertainment, Technology, Business, etc\}, where each centers could be composed of hidden clusters such as: \{Entertainment$|$ Star, Movie, TV play, etc\}; the corresponding cluster centers for year 2020 should be close to that in year 2010. 2) personalized concept correlation, \emph{i.e.,} the original meaning of past personalized concepts should be retained when learning new concepts.  For instance, the semantic meanings of \emph{Artificial Intelligence} for year 2010 and year 2020 should be very similar among consecutive news cluster tasks. Thus, the feature embedding or representation of \emph{Artificial Intelligence} for these two tasks should be identical accordingly.

\item  ``\textbf{Catastrophic Neglecting}'', \emph{i.e.,} how to correctly generate multiple personalized concepts from one prompt while avoiding the semantic catastrophic neglect of target prompt. For the correction when generating creative image with a text prompt, \emph{e.g.,} ``\textit{a photo of pet dog with James}'', how to overcome the influence of concept-neglecting is a common thought in text-to-image diffusion models. Moreover, another concerning phenomenon known as attribute-neglecting or catastrophic homogenization is also within the Stable Diffusion model. Specifically, when generating an image consisting of two objects sharing similar visual characteristics, exemplified by the prompt ``photo of a dog and cat", the Stable Diffusion model falls victim to attribute-neglecting, resulting in the generation of an image portraying either "two dogs" or "two cats".
    %For the efficiency when facing a new clustering task, how to overcome the influence of past learned tasks is a common thought in lifelong machine learning since accessing the previous data can be time-consuming. One common efficient method is to alternatively optimize the task-specific components and the common knowledge.
\end{itemize}

%through our empirical investigations, we have discovered another concerning phenomenon known as catastrophic homogenization within the Stable Diffusion model.
%Specifically, when tasked with generating an image consisting of two objects sharing similar visual characteristics, exemplified by the textual prompt "photo of a dog and cat," the Stable Diffusion model falls victim to homogenization, resulting in the generation of an image portraying either "two dogs" or "two cats".

After summarizing the aforementioned, here we propose a lifelong diffusion framework for continual personalized text-to-image diffusion model. Our model is memory efficient and computation effective when augmenting with a sequence of new concepts. To tackle the challenges in details, the underlying hypothesis is that the pre-trained diffusion model can be updated with a few samples of the new concept, which motivates we design a \underline{L}ife\underline{l}ong {t}ext-to-{i}mage \underline{D}iffusion \underline{M}odel (L$^2$DM) via considering lifelong-task learning and multi-concept generation. To be specific, we develop a task-aware memory enhancement module and a elastic concept distillation module for preserving prior knowledge and personalized knowledge, respectively.
These modules could safeguard the identifying features amongst prior concepts and learned personalized concepts in the lifelong generation task sequence. Meanwhile, we explore a novel concept attention artist module and an orthogonal attention module to respectively address the concept-neglecting and attribute-neglecting, which encourages all the concepts to attend in the image when generating multiple concepts. When observing a new concept or generation task provided via a user, the lifelong learning mechanism of our L$^2$DM can elastically learn concept knowledge with a rainbow-memory bank strategy. Then the attention artist modules could guide and further manifest the semantics information in the generated image with a given prompt. We carry out several extensive experiments to demonstrate the performance of our L$^2$DM model in comparison with the state-of-the-arts. The ablation studies also highlight the contribution of each component in the potential lifelong text-to-image diffusion challenge.

\indent The novelties of our proposed L$^2$DM model can be summarized as:
%\vspace{-3pt}
\begin{itemize}[leftmargin=12pt]
\setlength{\itemsep}{1pt}
\setlength{\parsep}{0pt}
\setlength{\parskip}{0pt}
  \item To the best of our knowledge, we in this paper take the first attempt in exploring a \underline{L}ife\underline{l}ong {t}ext-to-{i}mage \underline{D}iffusion \underline{M}odel, \emph{i.e.,} L$^2$DM, which can continually add and perform new text-to-image diffusion tasks via transferring past accumulated concept knowledge.
       %, this is the first attempt to study the problem of spectral clustering in the lifelong learning setting, i.e., Lifelong Spectral Clustering ($\mr{L^2SC}$), which can adopt previously accumulated experience to incorporate new cluster tasks, and improve the clustering performance accordingly.
  \item For ``catastrophic foregetting'' issue, we consider a task-aware memory enhancement module and an elastic concept distillation module in our  L$^2$DM, which can respectively consolidate the prior knowledge from rehearsal respect, and elastically distill the personalized knowledge amongst text-to-image diffusion models.
  %the  capture the hierarchical clustering centers and feature embedding correlations amongst consecutive clustering tasks.
         %can  We present two common knowledge libraries: orthogonal basis library and feature embedding libray, which can simultaneously preserve the latent clustering centers and capture the feature correlations among different clustering tasks, respectively.
  \item In respect of ``catastrophic neglecting'' issue, we present a novel concept attention artist module and an orthogonal attention module when generating multiple user-specific concepts, which could activate all the described semantics in the text prompt. Various experiment results strongly support the efficiency and effectiveness of our L$^2$DM.
        %We propose an alternating direction optimization algorithm to optimize the proposed $\mr{L^2SC}$ model efficiently, which can incorporate fresh knowledge gradually from online dictionary learning perspective. Various experiments demonstrate the effectiveness and efficiency of our proposed model.
       %With the advantage of separable structure in the proximal operation, we show that parallel block coordinate descent can optimize the proposed method in a efficient way.
\end{itemize}
We organize the rest of the paper as follows: the first section briefly review some related works. The second one introduces our proposed lifelong text-to-image diffusion problem with the L$^2$DM framework. Then, how to efficiently justify the proposed model is proposed in the experimental results, followed by the conclusion and limitations.

%Our work mainly draws from \textbf{Multi-task Clustering} and \textbf{Lifelong Learning}, so we give a brief review on these two topics.
\section{Related Work}\label{sec:related work}
This section briefly reviews several representative works on two related topics, \textbf{Lifelong Machine Learning} and \emph{i.e.,} \textbf{Text-to-image Diffusion}.
%Multi-task learning and its related methods have a long history. Depending on whether multi-task learning can be learned in an online way, the multi-task learning model can be divided into : \textbf{Batch Multi-task Learning} and \textbf{Lifelong Learning}.

\indent For the related \textbf{Lifelong Machine Learning}, the earlier works intends to transfer the selective information from task cluster to new tasks \cite{thrun1996discovering}, or transferring invariance knowledge by using neural networks \cite{thrun2012explanation}. The main target is to learn on non-stationary data streams without catastrophically forgetting previous knowledge. To achieve this, \cite{Sun9037204} proposes a new continual multi-view task learning model by combining sparse subspace learning and deep matrix factorization; \cite{ammar2015autonomous} presents a cross-domain lifelong learning framework for reinforcement learning, which can improve the efficiency when specializing to different domains. Different from the latent knowledge-based algorithms, ~\cite{kirkpatrick2017overcoming} develops an elastic weight consolidation (EWC) model, and implements it via using the diagonal of the Fisher information matrix; \cite{li2016learning} proposes a learning without forgetting (LwF) method into convolutional neural networks, which could retrain the neural networks without using the past task data. These regularization-based algorithms can consolidate the learned knowledge via constraining the neural network parameters. Instead of fixing the neural networks,~\cite{yoon2018lifelong} proposes a novel deep network to dynamically decide its network capacity when a sequence of tasks arrive; DER \cite{yan2021dynamically} adds a new learnable feature extractor for its network architecture while freezing the previously learned feature extractor, when learning each a new task; DyTox~\cite{douillard2022dytox} proposes a transformer architecture with dynamic expansion of task tokens in the continual learning paradigm. Amongst the recently-proposed discussions above, there is no works concerning how to extending lifelong learning in the text-to-image fields, and our work is a pioneering work to attain lifelong text-to-image generation in computer vision and robot perception applications.

%However, the model in \cite{sun2020lifelong} considers how to use linear task correlations among spectral tasks, while our current work employs neural networks to capture the hierarchical and nonlinear relationships among consecutive spectral tasks.
%Among the discussion above, there is no works concerning how to extend lifelong machine learning to unsupervised learning setting, e.g., spectral clustering, and our current work represents the first work to achieve lifelong spectral clustering learning.

%incorporates
For the \textbf{Text-to-Image diffusion}, diffusion models\cite{rombach2022high,song2020denoising, nichol2021improved} (DM) have demonstrated remarkable generative power in various fields, such as image-to-image generation\cite{saharia2022palette,wang2022pretraining}, text-to-image generation\cite{rombach2022high, saharia2022photorealistic}, text-to-video generation\cite{blattmann2023align,singer2022make} and text-to-3D generation\cite{poole2022dreamfusion,lin2023magic3d}. By encoding text inputs into a condition vector using a pretrained CLIP~\cite{radford2021learning}, text-to-image diffusion have achieved successful applications in image generation. As examples, GLIDE\cite{nichol2021glide} represents a text-guided diffusion model that encompasses both image generation and editing capabilities; Imagen\cite{saharia2022photorealistic} adopts classifier free guidance and a pretrained large language model for image generation. In particular, as a significant advancement of latent diffusion, Stable Diffusion\cite{rombach2022high} utilizes VQ-GAN\cite{esser2021taming} to achieve the mapping of the representation space from pixel space to latent space. This transformation can not only reduce the computational burden associated with processing large-scale data samples, but also enable the model to focus the comprehension of well-compressed semantic features and visual patterns. In parallel, both StructureDiffusion\cite{feng2022training} and Attend-and-Excite\cite{chefer2023attend} identify the risk of catastrophic neglect in synthetic images of pretrained Stable Diffusion, and mitigate the effects of this risk via introducing structured guidance or refining the cross-attention units.

\textbf{Personalized Text-to-Image diffusion:} In order to leverage the potential of Stable Diffusion\cite{rombach2022high} for synthesizing images that align with the user's own concepts, recent works have focused on the customization and personalization of text-to-image diffusion models through fine-tuning techniques applied to personalized datasets. For instance DreamBooth\cite{ruiz2023dreambooth} employs a class-specific prior preservation loss to safeguard prior knowledge and conducts fine-tuning of all parameters of Stable Diffusion. In contrast, Textual Inversion\cite{gal2022image} focuses on learning word vectors solely for new concepts. Custom Diffusion\cite{kumari2022multi} selectively fine-tunes a limited number of parameters within specific attention layers, facilitating fast tuning. Meanwhile, SVDiff\cite{SVDiff} leverages Singular Value Decomposition (SVD) on the weight matrices of Stable Diffusion, thereby enabling more efficient tuning on personalized datasets. However, all the above models cannot consecutively incorporate a new text-to-image diffusion task without reusing the past data.

%\textbf{Notations}
%For matrix $W\in \mathbb{R}^{m\times n}$, let $w_{ij}$ be the entry in the $i$-th row and $j$-th column of $W$. Let us define some norms, $\left\|W\right\|_0$ is the number of nonzero entries in $W$; denote by $\left\|W\right\|_1=\sum_{i=1}^m\sum_{j=1}^n|w_{ij}|$ and $\left\|W\right\|_{\infty}=\max_{i,j}|w_{ij}|$ the $\ell_1$-norm and $\ell_{\infty}$-norm of $W$, respectively. Let $\left\|W\right\|_{2,1}=\sum_{i=1}^m\left\|w_i\right\|_2$; denote by $\mr{sgn}$, $\odot$ and $\lvert \cdot\lvert$ the elementwise sign, multiplication and absolute value of matrix, respectively. Let $(\cdot)_+$ be the positive part elementwise of matrix.

% let $\left\|W\right\|_*=\sum_{i=1}^r \sigma_i(W)$ be the trace norm of $W$, where $r=\mr{rank}(W)$ and $\sigma_i(W)$ be the singular values of $W$;

%The key insight of LSC is that adopting previously learned spectral clustering tasks can improve the clustering performance of future spectral tasks and reduce learning time accordingly, as new spectral clustering task is coming gradually.  Therefore in the rest of this section,
\begin{figure*}[t]
	\centering
	\includegraphics[width=515pt]
	     {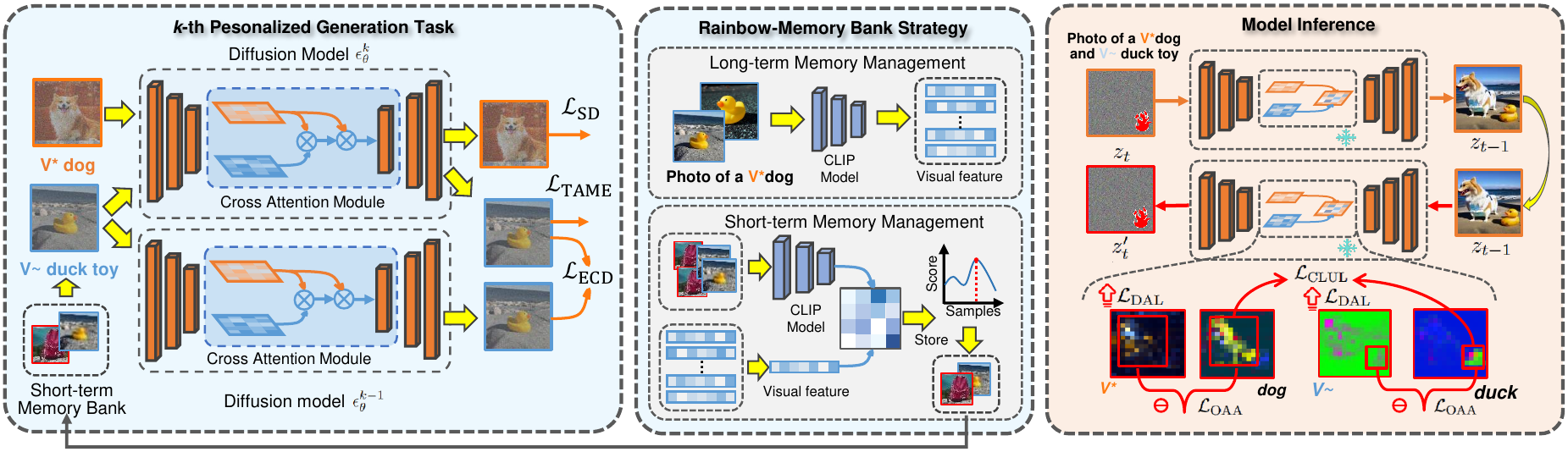}
        \caption{Overview of our proposed framework, where the new concept \textit{V* dog} with a few examples can be added faster in our lifelong text-to-image diffusion model, and the $\mc{L}_{\mr{SD}}$, $\mc{L}_{\mr{TAME}}$ and $\mc{L}_{\mr{ECD}}$ denotes the Stable Diffusion model loss, task-aware memory enhancement loss, and elastic concept distillation loss, respectively. The diffusion model inference stage can be efficiently achieved with a concept attention artist ($\mc{L}_{\mr{CLUL}}$ and $\mc{L}_{\mr{DAL}}$) and an orthogonal attention artist module ($\mc{L}_{\mr{OAA}}$).}
	\label{fig: overview}
	\vspace{-10pt}
\end{figure*}

\section{Lifelong Text-to-Image Diffusion Model (L$^2$DM)} \label{sec: Method}
In this section, we will introduce our proposed lifelong text-to-image diffusion method from several aspects. We first present some background on text-image diffusion model in Sec.~\ref{subsec:revisit}, and the problem definition for lifelong diffusion framework in Sec.~\ref{subsec:problem}. Moreover, we propose our model in Sec.~\ref{subsec:our_framework}, which consists of how to attain continual concept learning and multiple concepts generation. %followed by the object attention artist for lifelong multi-concept generation.

%$\mathcal{D}$ is a latent decoder
\subsection{Revisit Text-to-Image Diffusion Model}\label{subsec:revisit}
Diffusion models (\emph{e.g.,} Stable Diffusion\cite{rombach2022high}) rely on a sequential generation process that gradually refines the generated image over multiple iterations, and aims to bridge the semantic gap between text and images by learning a mapping from textual descriptions to visual representations. Relying on a pair of pre-trained latent compression models consisting of a latent encoder $\mathcal{E}$ and a decoder $\mathcal{D}$, latent diffusion models (\emph{i.e.,} LDMs)\cite{rombach2022high} can directly predict a representation in a low-dimensional and enriched-semantic latent space. To be specific, we in this paper apply a pre-trained Stable Diffusion model to perform text-to-image generation task, which is a variant of LDMs. Given a text prompt $p$ and an initial noise map $\varepsilon \in \mathcal{N}(0,\bm{I})$, an image $x$ can be generated by $x=\mathcal{D}(\epsilon_\theta(c,t))$, where $\epsilon_\theta$ denotes the diffusion model, $c=\mc{P}(p)$ and $\mc{P}$ is a text encoder from CLIP~\cite{ramesh2022hierarchical}. Formally, the diffusion model $\epsilon_\theta$ is trained to denoise the noise image and predict a denoised output as follows:
\begin{equation} \label{eq: ldm}
\begin{aligned}
      \mc{L}_{\mr{LDM}}(\theta) := \mathbb{E}_{z,c,\varepsilon,t}\big[\Vert \varepsilon - \epsilon_\theta(z_{t}|c,t)\Vert^2_2\big],
\end{aligned}
\end{equation}
where $(z_t,c)$ are the corresponding pairs of image latents and text embeddings, $z_t$ denotes a latent code obtained from encoder $\mathcal{E}$, and $\theta$ denotes the corresponding model parameters. $t \in \mr{Uniform}(1, T)$, which will be omitted in the following sections for brevity.

However, recent works (\emph{e.g.,}~\cite{ruiz2023dreambooth}) have focused on personalizing text-to-image models via a fine-tuning manner. For instance, a pre-trained text-to-image diffusion model can be fine-tuned on a personalized dataset consists of $3\sim5$ images paired with their text description. In light of this, the \underline{P}ersonalized text-to-image \underline{D}iffusion \underline{M}odel (\emph{i.e.,} PDM) can synthesize unprecedented images of the target concept with equivalent text prompts, where the prior knowledge should keep unchanged in this personalized model. To achieve this, DreamBooth\cite{ruiz2023dreambooth} proposes a class-specific prior preservation loss to resist overfitting for new concept, and protect the prior knowledge. The training objective of PDM becomes:
\begin{equation} \label{eq: pdm}
\begin{aligned}
      \mc{L}_{\mr{PDM}}(\theta) := & \mathbb{E}_{z,c,\varepsilon,t}\big[\Vert \varepsilon - \epsilon_\theta(z_{t}|c)\Vert^2_2\big]     \\
       \quad + & \lambda\mathbb{E}_{z^p,c^p,\varepsilon,t}\big[\Vert \varepsilon - \epsilon_\theta(z^p_{t}|c^p)\Vert^2_2\big],
\end{aligned}
\end{equation}
where $z_t^p$ is latent code encoded from generated image by Stable Diffusion, $c^p$ is the corresponding text condition, and $\lambda$ is the hyperparameter to control the backpropagation of the second term. However, this personalized text-to-image diffusion scenario does not consider continually incorporating new concepts without repeatedly using the past training data.

%\textbf{Personalized Text-to-Image diffusion models}. To obtain a personalized Text-to-Image diffusion model (PDM), a pre-traind Text-to-Image diffusion model is fine-tuned on a personalized dataset consists of 3$\sim$5 images paired with their text description. In light of this, the PDM can synthesize unprecedented images of the target concept with equivalent text prompts. At the same time, the prior knowledge should keep unchanged in the personalized model. For this, Dreambooth proposes a class-specific prior preservation loss to resist overfitting for new concept and protect the prior knowledge.

\subsection{Problem Definition for Lifelong Text-to-Image Diffusion}\label{subsec:problem}
Suppose that one user wishes to continually synthesize new specific concepts over a large-scale diffusion model, where these generation tasks are defined as $\mathcal{T} = \{\mathcal{T}^k\}_{k=1}^K$ with datasets $\{\mathcal{D}^k\}_{k=1}^K$, and the $k$-th generation task consists of $n^k$ pairs of concept image $x^k_i$ and prompt $c^k_i$, where $K$ denotes the total task quantity, and $n^k$ always is $3\sim5$ in this paper. Different from the recent prior-based personalized generation method, we here consider the scenario that a text-to-image diffusion model encounters a series of consecutive specific concepts over a lifelong time, where each concept generation task $\mc{T}^{k}$ can be expressed as Eq.~\eqref{eq: pdm}. For each timestamp, as the $n^k$ pairs of concept image $x^k_i$ and prompt $c^k_i$ for the task $k$ are imposed into the lifelong text-to-image diffusion by user, this system should incorporate the new concepts while composing new concepts with learned concepts efficiently.  Without considering  privacy leakage and limited memory, we assume that the data pairs are only available for the current task, \emph{i.e.,} $\cup_{i=1}^{n^k}(x^k_i,c^k_i) \cap(\cup_{\ell=1}^{k-1}\cup_{i=1}^{n^{\ell}} (x^{\ell}_i,c^{\ell}_i))=\emptyset$, and the training objective function for lifelong text-to-image diffusion problem can be formulated as:
\begin{equation} \label{eq: eq3}
\begin{aligned}
      \mc{L}_{\mr{L^2DM}} (\theta) : &= \! \sum^K_{k=1}  \Big\{\mathbb{E}_{z^k,c^k,\varepsilon,t}\big[\Vert \varepsilon - \epsilon^k_\theta(z^k_{t}|c^k)\Vert^2_2\big] +\lambda \mc{L}^ {k}_{\mr{PR}}(\theta) \Big\}  \\
       \mc{L}^{k}_{\mr{PR}} (\theta): &=\mathbb{E}_{z^{k,p},c^{k,p},\varepsilon,t}\big[\Vert \varepsilon - \epsilon^k_\theta(z^{k,p}_t|c^{k,p})\Vert^2_2\big],
\end{aligned}
\end{equation}
where $(z_t^k,c^k)$ are the corresponding pairs of image latents and text embeddings for the $k$-th generation task, and $(z_t^{k,p},c^{k,p})$ denotes the prior pairs generated by the pre-trained model for the $k$-th generation task. More specifically, the goal in the above equation is to obtain the diffusion model parameter $\theta$ with the target as follows:
\begin{itemize}
    \item Generation Performance: the lifelong diffusion model could accurately and correctly generate multiple personalized concepts without drifting or neglecting from the original generalization ability;
  \item Computation Efficiency: the new text-to-image diffusion task or model should be added faster than conventional personalized text-to-image diffusion model, when facing a new concept or generation task provided by user;
   \item Lifelong Learning: the lifelong learning system could learn a new concept or generation task arbitrarily and efficiently when the user provides the personal concept to synthesize new variations.
\end{itemize}
To attain ``lifelong learning'' with acceptable generation performance and computation efficiency, we introduce our lifelong text-to-image diffusion model from perspectives of tackling ``catastrophic forgetting'' and ``catastrophic neglecting'' issues as following.

\subsection{The Proposed L$^2$DM Framework}\label{subsec:our_framework}
As shown in Fig.~\ref{fig: overview}, we present a pioneering \underline{L}ife\underline{l}ong text-to-image \underline{D}iffusion \underline{M}odel (\emph{i.e.,} L$^2$DM), which aims at continually learning new personalized generation tasks while retaining prior knowledge. To specifically address the issue of catastrophic forgetting concerning lifelong-task learning, we devise a {t}ask-{a}ware {m}emory {e}nhancement module (TAME) and an {e}lastic {c}oncept  {d}istillation (\emph{i.e.,} ECD) module. These two modules work in tandem to safeguard the knowledge of previous tasks and expand the representational capacity for newly encountered task in an effective manner. To tackle catastrophic neglect and homogenization during multi-concept inference stage of our L$^2$DM model, we further design an concept attention artist (CAA) module and an orthogonal attention artist (OAA) module. These two modules serves to safeguard the representation of each individual concept, mitigating the risk of catastrophic neglecting of identifying features and preventing it from being detrimentally contaminated by other concepts.

%homogenization
%One natural idea is to retain or fine-tune the diffusion model via storing all the user-specific concepts, which could inevitably cause enormous computation and storage.
%during the fine-tuning process on the target dataset, the concept imbalance between the source and target datasets can lead the model to overfit to the target dataset, resulting in subpar performance on the source dataset.
%Furthermore, we put forth a \underline{R}ainbow \underline{M}emory \underline{B}ank (\emph{i.e.,} RMB) that stores long-term memory and short-term memory to optimize the functionality of the task knowledge revising module during the learning of new task.

%To explore the dependencies among different generation tasks, one straightforward way is extending the classical lifelong learning methods (\emph{i.e.,} LWF~\cite{li2017learning} and EWC~\cite{kirkpatrick2017overcoming}) into text-to-image diffusion model. However, these methods above will significantly overfit on the new tasks, while causing past concepts omission when composing multiple concepts.
\subsubsection{Lifelong-task Learning}
In the lifelong text-to-image diffusion setting, one major concern is how to overcome the catastrophic forgetting problem, when the knowledge in a large-scale diffusion model is transferred from learned tasks to a new task. Different from conventional lifelong learning problem (\emph{i.e.,} LWF~\cite{li2017learning} and EWC~\cite{kirkpatrick2017overcoming}), we in this work consider two distinct forgetting aspects within our L$^2$DM framework: 1) {prior-forgetting}, integrating new personalized concepts into a large text-to-image diffusion model could forget the learned prior concepts; 2) {personalized-forgetting}, as our L$^2$DM continually learns user personalized concepts, which could lead to the catastrophic forgetting of previously learned personalized concepts. To tackle these above challenges of lifelong-task learning, we in the following section describe each component of our lifelong learning algorithm in details: 1) task-aware memory enhancement, and 2) elastic concept distillation.

%one of the reasonable manners is to jointly optimize the diffusion model by constraining the datasets across prior concepts and personalized concepts. This strategy can effectively preserve the knowledge of prior concepts and mitigates the potential risk of model overfitting. However, the method cannot be applied to tackle the issue of both forgetting issues, since the prior data could not be obtained with the limitation of privacy.

\begin{figure}[t]
	\centering
	\includegraphics[width=248pt, height=135pt]
	     {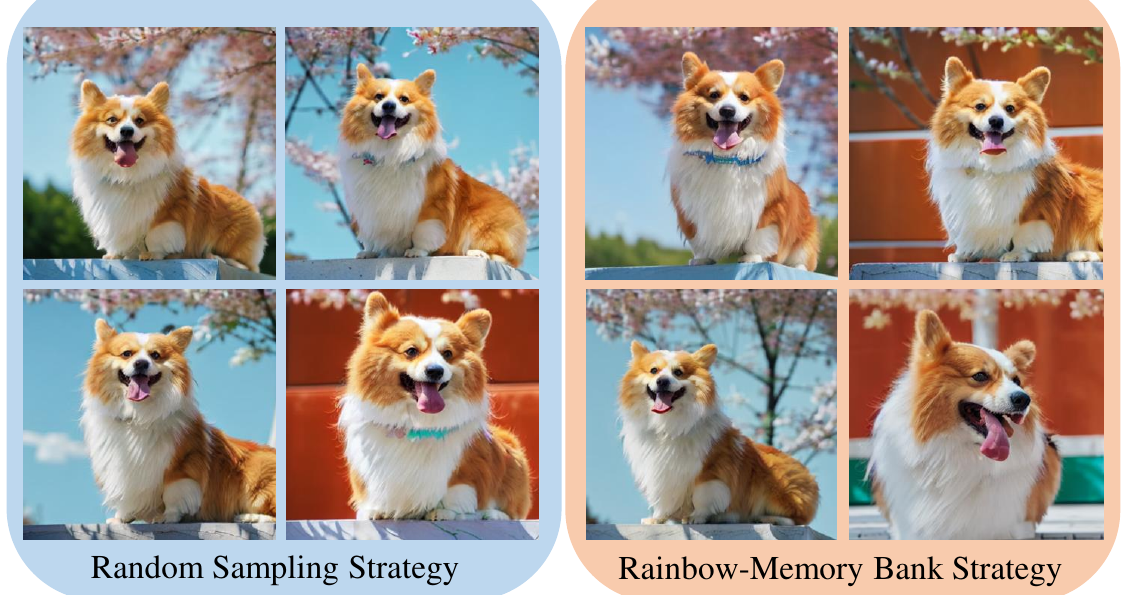}
        \caption{Comparison results between random sampling strategy with our rainbow-memory bank strategy, where our strategy can select more discriminative images in terms of views and poses for the personalized concept ``dog''.}
	\label{fig: rss}
	\vspace{-10pt}
\end{figure}

%In the presented framework in Eq.~\ref{eq:our_tame}, we propose two additional terms to effectively tackle the two distinct aspects of forgetting within the lifelong text-to-image diffusion model.  that arises during the joint training of multiple concepts, we first expand $\mc{L}^{k}_{\mr{PR}}(\theta)$ to derive the third term of Eq.~\ref{eq:our_tame}   To partially mitigate the personalized-forgetting issue, the second term of Eq~\eqref{eq:our_tame} is applied to provide supervision to our L$^2$DM using a short-term memory, which is composed of the generated data corresponding to the learned personalized concepts. This term enables the model to actively revise its prior knowledge, and strengthen its memory retention of the learned personalized concepts. The details about how to select our short-term memory and store long-term memory are described as below:
%\textcolor[rgb]{1.00,0.00,0.00}{Then the short-term memory bank will be stored and recast in a long-term memory.}
\textbf{Task-Aware Memory Enhancement (TAME):} To tackle {prior-forgetting} when facing with a new user-specific concept, one naive approach is to retain or fine-tune the diffusion model via storing dataset of user-specific concepts. However, this strategy could inevitably cause enormous computation and storage cost, when encountering large-scale user concepts. Additionally, the training data could not be easily accessed due to the privacy limitation. Inspired by Eq~\eqref{eq: pdm}, we develop a task-aware memory enhancement module to address prior-forgetting from rehearsal mechanism while partially preserving the personalized-forgetting. The memory enhancement module for our L$^2$DM can be formulated as:
\begin{equation}
\begin{aligned}\label{eq:our_tame}
       \mc{L}_{\mr{TAME}} (\theta) &:= \alpha\sum^{k-1}_{\ell=1}\mathbb{E}_{\hat{z}^{\ell},\hat{c}^{\ell},\varepsilon,t}\big[\Vert \varepsilon - \epsilon^k_\theta(\hat{z}^{\ell}_{t}|\hat{c}^{\ell})\Vert^2_2\big]   \\ &+\beta \sum^{k}_{\ell=1}\mc{L}^{k}_{\mr{PR}}(\theta),
      \end{aligned}
\end{equation}
where \ $\hat{}$ \ represents the input $\hat{x}^{\ell}$ is from short-term memory, and the hyperparameter $\mu$ is utilized to reweight the loss function. To be specific, the first term of Eq~\eqref{eq:our_tame} is applied to provide supervision for our L$^2$DM using a fine-grained memory bank, which is composed of the generated data corresponding to the learned personalized concepts. In this way, this term enables the diffusion model to actively preserve its prior knowledge, and strengthen its memory retention of the learned personalized concepts. The details about how to select our short-term memory and store long-term memory are described in \textbf{Rainbow-Memory Bank Strategy}. Driven by the DreamBooth\cite{ruiz2023dreambooth}, the second term of Eq.~\ref{eq:our_tame} is derived via expanding $\mc{L}^{k}_{\mr{PR}}(\theta)$, which could reduce the impact of prior-forgetting issue. Meanwhile, this term could promote output diversity within our L$^2$DM model, and mitigate the impact of uninterrupted personalized concepts on the pre-trained Stable Diffusion model.
\renewcommand{\algorithmicrequire}{\textbf{Input:}}
\renewcommand{\algorithmicensure}{\textbf{Output:}}
\begin{algorithm}[t]			
	\caption{Rainbow-memory selection strategy}
	\label{alg1}
    \setlength{\tabcolsep}{1.2mm}
	\begin{algorithmic}[1]
		\REQUIRE Image encoder $\mc{I}$, diffusion model $\epsilon^k_\theta$, short-term memory $\mc{B}^k_s$, long-term memory bank $\mc{B}^{1:k-1}_l$, hyper-parameter $\eta$, current dataset $\mc{D}^k$;
         \ENSURE Memory bank $\{\mc{B}^k_s,\mc{B}^{1:k}_l\}$;
        \STATE \textbf{\#Short-term Memory Update\#}
        \IF{$k>1$}
		\FOR {$\ell=1, 2, \cdots, k-1$}
            \STATE Generate images $\{\hat{x}_i^{\ell}\}^{\eta}_{i=1}$ for each $p^{\ell}_i$ in $\mc{B}^{1:k-1}_l$;   
        		\ENDFOR
            \FOR {$ f_l^\ell \  \mathrm{in} \ \mc{B}^{1:k-1}_l$}
            \FOR {$ \hat{x}_a^{\ell}  \  \mathrm{in} \ \{\hat{x}_a^{\ell}\}^{\eta}_{a=1}$}
            \STATE Compute image feature $\hat{f}^{\ell}_a$ by $\mc{I}$;
            \STATE Compute score $S(\hat{f}_a^{\ell}, f^\ell_l)$ using Eq.~\ref{eq: eq5};
            \ENDFOR
            \STATE Store $\hat{x}_a^{\ell}$ with the highest $S(\hat{f}_a^{\ell}, f^\ell_l)$ in $\mc{B}^k_s$;

           \ENDFOR 
           \ENDIF
          \RETURN $\mc{B}^k_s$ \\
        \STATE \textbf{\#Long-term Memory Update\#}
          \FOR{$(x^k_i, p^k_i) \ \mr{in} \ \mc{D}^k$}
          \STATE Compute image feature $f^k_i$ by $\mc{I}$;
          \STATE Store $f^k_i$ and task prompt $p^k_i$ in $\mc{B}_l^{1:k-1}$;
          \ENDFOR
          \STATE Initialize $\mc{B}_l^{1:k}$ with $\mc{B}_l^{1:k-1}$;
           \RETURN $\mc{B}_l^{1:k}$ \\
        
           \end{algorithmic}
\end{algorithm}

$\bullet$ \textbf{Rainbow-Memory Bank Strategy}: To establish a short-term memory, we need to select several samples from generated data during the optimization process of our L$^2$DM using Eq.~\eqref{eq:our_tame}. However, as shown in DreamBooth\cite{ruiz2023dreambooth}, there is a potential risk of diminishing output variability in terms of concept poses and views when fine-tuning on a few-shot personalized dataset, as the example depicted in Fig.~\ref{fig: rss}. Additionally, the generated data could be contaminated by some elements from the earlier learned concepts. To tackle the aforementioned issues and select images with high quality for our TAME module, a rainbow-memory bank strategy is designed with a short-term memory bank $\mc{B}_s^{k}$ and a long-term memory bank $\mc{B}_l^{1:k}$ for the current $k$-th task. In details, when learning the current $k$-th generation task, the short-term memory bank $\mc{B}_s^{k}$ is built from CLIP feature space with the guidance of the $k-1$-th long-term memory bank $\mathcal{B}_l^{1:k-1}$, where $\mathcal{B}_l^{1:k-1}$ accumulates the CLIP feature distribution $\cup_{\ell=1}^{k-1}\cup_{i=1}^{n^{\ell}}f_i^{\ell}$ and corresponding prompts $\cup_{\ell=1}^{k-1}\cup_{i=1}^{n^{\ell}}p^{\ell}_i$ amongst past generation tasks, \emph{i.e.,} $\mathcal{B}_l^{1:k-1}\leftarrow\cup_{\ell=1}^{k-1}\cup_{i=1}^{n^{\ell}}(f_i^{\ell},p^{\ell}_i)$. In this way, the short-term memory bank $\mc{B}_s^{k}$ can be established with the CLIP features $\cup_{\ell=1}^{k-1}\cup_{a=1}^{n^{\ell}}\hat{f}_a^{\ell}$, which corresponds to the generated images $\cup_{\ell=1}^{k-1}\cup_{a=1}^{n^{\ell}}\hat{x}_a^{\ell}$ in $\mc{B}^{1:k-1}_l$ using current diffusion model $\epsilon^k_\theta$. Considering the identifying characteristics for each past generation task, we define a score function to further cast the short-term memory bank $\mathcal{B}_s^{k}$: 
\begin{equation}\label{eq: eq5}
\begin{aligned}
  S(\hat{f}^{\ell}_a,f^\ell_i) &=\frac{\sum_{{\ell}'=1}^{k-1}\sum_{b=1}^{N^{{\ell}'}_g}1-\mr{sim}(\mc{I}(\hat{f}^{\ell}_a),\mc{I}(\hat{f}^{{\ell}'}_{b}))
  \!}{\sum^{k-1}_{{\ell}'=1}N^{{\ell}'}_g} 
 \\ & \quad+ \beta \mr{sim}(\mc{I}(\hat{f}^{\ell}_a),\mc{I}(f^\ell_i)),
\end{aligned}
\end{equation}            
where $\hat{f}^{\ell}_a$ and $\hat{f}^{{\ell}'}_{b}$ denote the corresponding CLIP features for the $a$-th generated sample of $\ell$-th task and the $b$-th generated sample of ${\ell'}$-th task, feature $f^{\ell}_i$ corresponds to the $i$-th sample of $\ell$-th task in the bank $\mc{B}_l^{1:k}$, $\mr{sim}(\cdot)$ denotes the cosine similarity, $N^{{\ell}'}_g$ is the total number of generated images for ${\ell'}$-th learned task and the $\beta>0$ is a trading-off parameter. The first term in Eq.~\eqref{eq: eq5} is designed to assess the catastrophic forgetting degree in generated images; the second term lies in its pivotal role of introducing the guidance of long-term memory bank. Then the $\hat{f}^\ell_a$ with a higher score will be stored to obtain the short-term memory bank $\mc{B}_s^{k}$, and the long-term memory bank $\mathcal{B}_l^{1:k}$ will be recast as $\mathcal{B}_l^{1:k-1}\cup_{i=1}^{n^{k}}(f_i^{k},p^{k}_i)$. We summarize the details for \textbf{Rainbow-Memory Bank Strategy} in \textbf{Algorithm}.\ref{alg1}.

\textbf{Elastic Concept Distillation (ECD):} Although the TAME module can enhance memory of the prior tasks or concepts, it only transfers the limited knowledge from the static images due to the limited of short-term memory bank $\mc{B}_s^k$. To further address personalized-forgetting issue, we consider the knowledge distillation technique to dynamically transfer diffusion knowledge from the last task. Different from the recent work~\cite{meng2023distillation} that focuses on trading-off between sample quality and diversity, we intend to maintain semantic consistency between the current model and last model. Therefore, current diffusion model $\hat{\epsilon}^k_\theta$ can be regarded as a student model to transfer knowledge form the last one $\hat{\epsilon}^{k-1}_\theta$. We then optimize the current model using the following objective:
\begin{equation} \label{eq: distillation}
\begin{aligned}
      \mc{L}_{\mr{ECD}}(\theta) &:= \gamma\sum^{k-1}_{\ell=1}\mathbb{E}_{\hat{z}^\ell,\hat{c}^\ell,\varepsilon,t}\big[\Vert \epsilon^{k-1}_\theta(\hat{z}^\ell_{t}|\hat{c}^\ell)- \epsilon^{k}_\theta(\hat{z}^\ell_{t}|\hat{c}^\ell)\Vert^2_2\big],
% \hat{\varepsilon}&:=\epsilon^{k-1}_\theta(\hat{z}^\ell_{t}|\hat{c}^\ell),
\end{aligned}
\end{equation}
where the latent code $\hat{z}^\ell_{t}$ and the corresponding text prompt $\hat{c}^\ell$ are from the short-term memory bank $\mc{B}_s^k$. Compared with TAME module, the ECD module can obtain a dynamic supervision map $\hat{\epsilon}^{k-1}_\theta(\hat{z}^\ell_{t}|\hat{c}^\ell)$ with the training iteration $t$, which can efficiently transfer semantic knowledge to tackle the personalized-forgetting issue.

%\textcolor[rgb]{1.00,0.00,0.00}{The input images and corresponding text prompts are then obtained from flash memory.} 

In conclusion, we first propose the TAME module that provides a strong rehearsal information to tackle prior-forgetting while partially respecting personalized-forgetting. Furthermore, ECD module is developed to obtain a soft guidance to transfer knowledge from the old diffusion model. The optimization of our L$^2$DM model can be simplified as:
\begin{equation}
\begin{aligned}\label{eq:our_l2dm}
       \mc{L}_{\mr{L^2DM}} (\theta) :&= \sum^K_{k=1}  \big\{\mathbb{E}_{z^k,c^k,\varepsilon,t}\big[\Vert \varepsilon - \epsilon^k_\theta(z^k_{t}|c^k)\Vert^2_2\big]
       \\
       &\quad +\mc{L}_{\mr{TAME}} (\theta)+ \mc{L}_{\mr{ECD}}(\theta)\big\}.
      \end{aligned}
\end{equation}

\subsubsection{Multi-concept Generation}
As aforementioned before, another core of lifelong text-to-image diffusion problem is the multi-concept generation problem, since the user wishes to compose multiple personalized concepts together. As present in Attend-and-Excite \cite{chefer2023attend}, an inherent issue in multi-concept generation is the \textbf{catastrophic neglecting}, which involves two key issues: 1) concept-neglecting, \emph{i.e.,} one or more user concepts of the input prompt cannot generated; 2) attribute-neglecting, \emph{i.e.,} tow objects share similar visual attributes or characteristics in a given prompt may fall victim to homogenization. To alleviate the issues above, a trivial solution for multi-concept generation is to dominant each concept token in some patches in the generated image (\emph{e.g.,} Attend-and-Excite~\cite{chefer2023attend}), or restrict the weight update to cross-attention key and value parameter (\emph{e.g.,} \cite{kumari2022multi}). {However, these methods just consider enhancing the representation of target texts, and ignore the interaction amongst multiple concepts}. To attain this, we explore a novel Concept {A}ttention {A}rtist (CAA) module for concept-neglecting issue, and present a {O}rthogonal {A}ttention {A}rtist (OAA) for attribute--neglecting issue.

\renewcommand{\algorithmicrequire}{\textbf{Input:}}
\renewcommand{\algorithmicensure}{\textbf{Output:}}
\begin{algorithm}[t]
\label{alg2}
\caption{Optimization Pipeline for Our Lifelong Text-to-image Diffusion Model}
\begin{algorithmic}[1]
\REQUIRE Personalized generation tasks $\{\mc{T}^k\}_{k=1}^K$ with datasets $\{\mc{D}^k\}_{k=1}^K$; \\
Pre-trained diffusion model: $\epsilon^0_\theta$;\\
Initialized memory bank: $\mc{B}^1_l=\emptyset$; \\
Text prompt $P$ with token indices $I_p$ and concept indices $I_c$; Timestep Number $T$,  Hyper-parameter $N_p$; \\
\ENSURE  $\epsilon^k_\theta$, $\mc{B}_l^{1:k}$;
\STATE \textbf{\#While observing a new task $\mc{T}^k$}:
\STATE \quad\textbf{\#Model Training\#}:
\STATE \quad Initialize short-term memory bank $\mc{B}^k_s=\emptyset$;
\STATE  \quad Store $\epsilon^{k-1}_\theta$ to initialize model $\epsilon^k_\theta$ and perform Eq.~\eqref{eq: distillation};
\STATE  \quad Update short-term memory bank $\mc{B}^k_s$ by Algorithm.~\ref{alg1};
\STATE \quad Generate $N_p$ prior images to build prior dataset $\mc{D}^p$;
\STATE \quad  \textbf{if} $k$=1 \textbf{then}
\STATE \quad  \quad Optimize model $\epsilon^k_\theta$ using Eq.~\eqref{eq: pdm} with $\mc{D}^k$, $\mc{D}^p$;
\STATE \quad \textbf{else}
\STATE \quad \quad Optimize model $\epsilon^k_\theta$ using Eq.~\eqref{eq:our_l2dm} with $\mc{D}^k$, $\mc{D}^p$, $\mc{B}_s^K$;
\STATE \quad  \textbf{end if}
\STATE \quad Update long-term memory bank $\mc{B}^{1:k}_l$ by Algorithm.~\ref{alg1}; \\

\STATE \quad \textbf{return} $\epsilon^k_\theta$; \\
\STATE \quad \textbf{\#Model Inference\#}:
\STATE  \quad  Initialize noise map $\varepsilon$ as $z_T$;
\STATE \quad \textbf{for} $t=T,T-1,\ldots,1$ \textbf{do};
\STATE \quad \quad Obtain attention map $A_t$ by $\epsilon^k_\theta(z_{t}|P)$;
\STATE \quad \quad Compute loss $\mc{L}_{\mr{CAA}}$ and $\mc{L}_{\mr{OAA}}$ using $A_t[:,:,I_p]$ and \\
\quad \quad $A_t[:,:,I_c]$; 
\STATE \quad \quad Update $z_t$ to $z_t'$ using back propagation;
\STATE \quad \quad $z_{t-1} \xleftarrow{} \epsilon^k_\theta(z_{t}'|P)$; 
\STATE \quad \textbf{end for}
\STATE \quad \textbf{return} $z_1$.

%\STATE Initialize $L_0$ in the first coming task
%\STATE Return $\{s_t\},\mc{D}$;
\end{algorithmic}
\end{algorithm}

\textbf{Concept Attention Artist (CAA):} considering the {concept-neglecting} issue, our empirical findings indicate that it suffers a significant influence of cross-attention representation amongst objects. Specifically, the spatial proximity of two concepts or a significant representation disparity of two concepts can lead to a consistent representation displacement of one concept by another one. In other words, when two concepts are closely positioned or when the representation values of one concepts significantly outweighs that of the other, the representation of one concepts tends to overshadow while further replacing the representation of the other one. To handle with these two issues, we develop a {concepts-level unmix loss} and a {dynamic attend loss} in our concept attention artist module to overcome the spatial proximity and representation disparity, respectively. The {concepts-level unmix loss} is designed to limit representation region of every attended concepts, which can be simplified as:   
\begin{align}
\label{eq: eq7}
     \mc{L}_{\mr{CLUL}} =\frac{1}{n_ln_c}\sum^{n_l}_{l=1} \sum^{n_c}_{i=1}\Vert A^i_{t,l} \cdot M_i\Vert^2_2,
\end{align}
where $n_c$ denotes the number of target tokens (\emph{e.g.}, `dog', `cat'), $n_l$ represents the quantity of the selected layers.  $A^i_{t,l} \in\mathbb{R}^{s\times n}$ denotes attention score of $i$-th token in the cross-attention module of diffusion model, and $l$ represents the $l$-th layer, and $s$ is the number of input tokens in $c$. $M_i$ represents a mask utilized in this loss function, which selectively assigns a value of zero to half of the cross-attention map while assigning a value of one to the remaining half. After leveraging the concept-level unmix loss, we are empowered to effectively disentangle individual concepts from the scene, thus circumventing the inherent mutual influence existed among concept representations in cross-attention map.

\begin{figure*}[htbp]
	\centering
	\includegraphics[width=500pt, height=615pt]
	{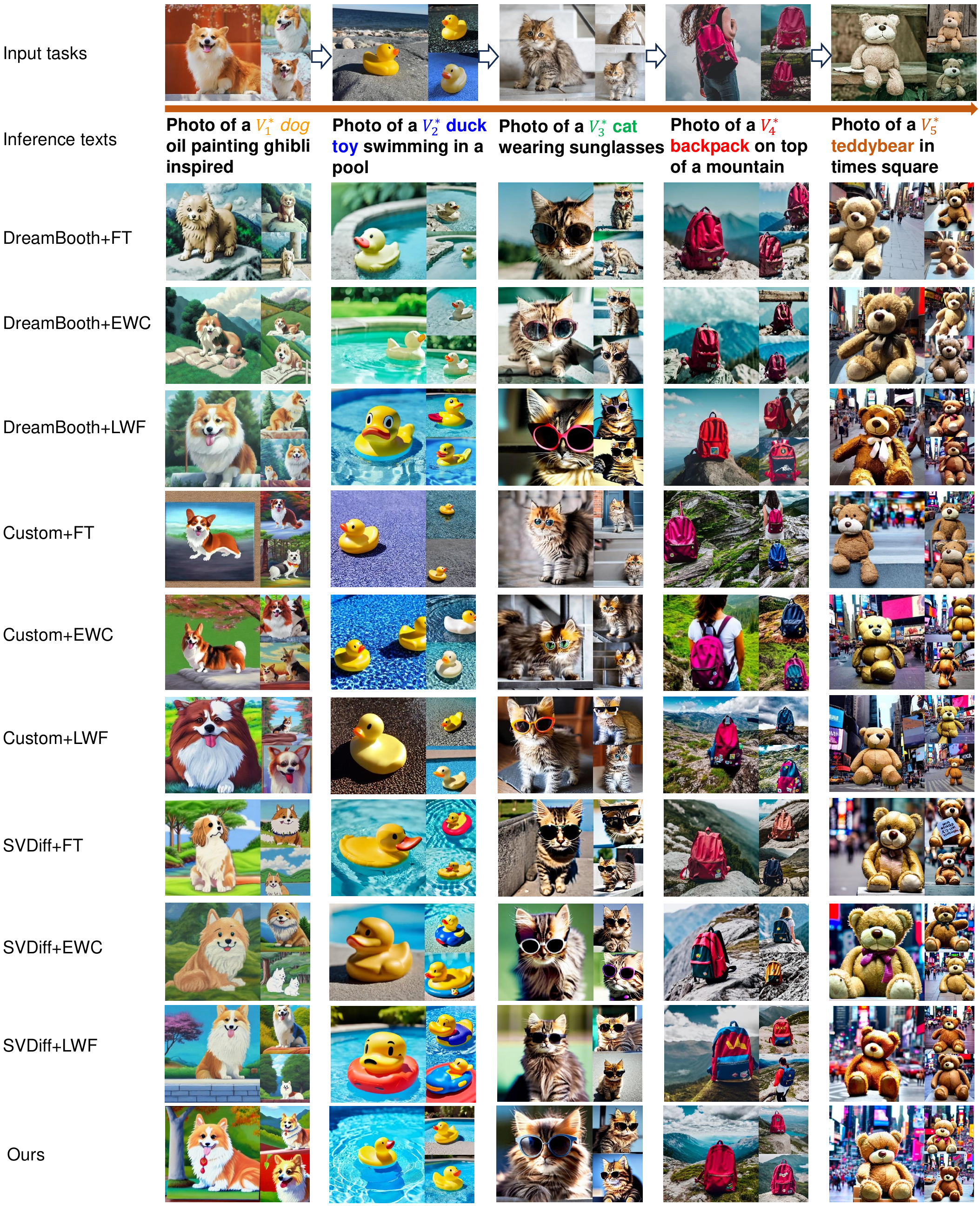}
        \caption{Qualitative comparisons between ours with the state-of-the-arts in the lifelong single-concept generation setting, where the first two rows denote the continual user-specific generation tasks needed by the user, and the rest rows denotes the generation results among the competing methods for each same prompt.}
	\label{fig: experiments_single}
	\vspace{-1pt}
\end{figure*}

\begin{figure*}[htbp]
	\centering
	\includegraphics[width=500pt, height=615pt]
	{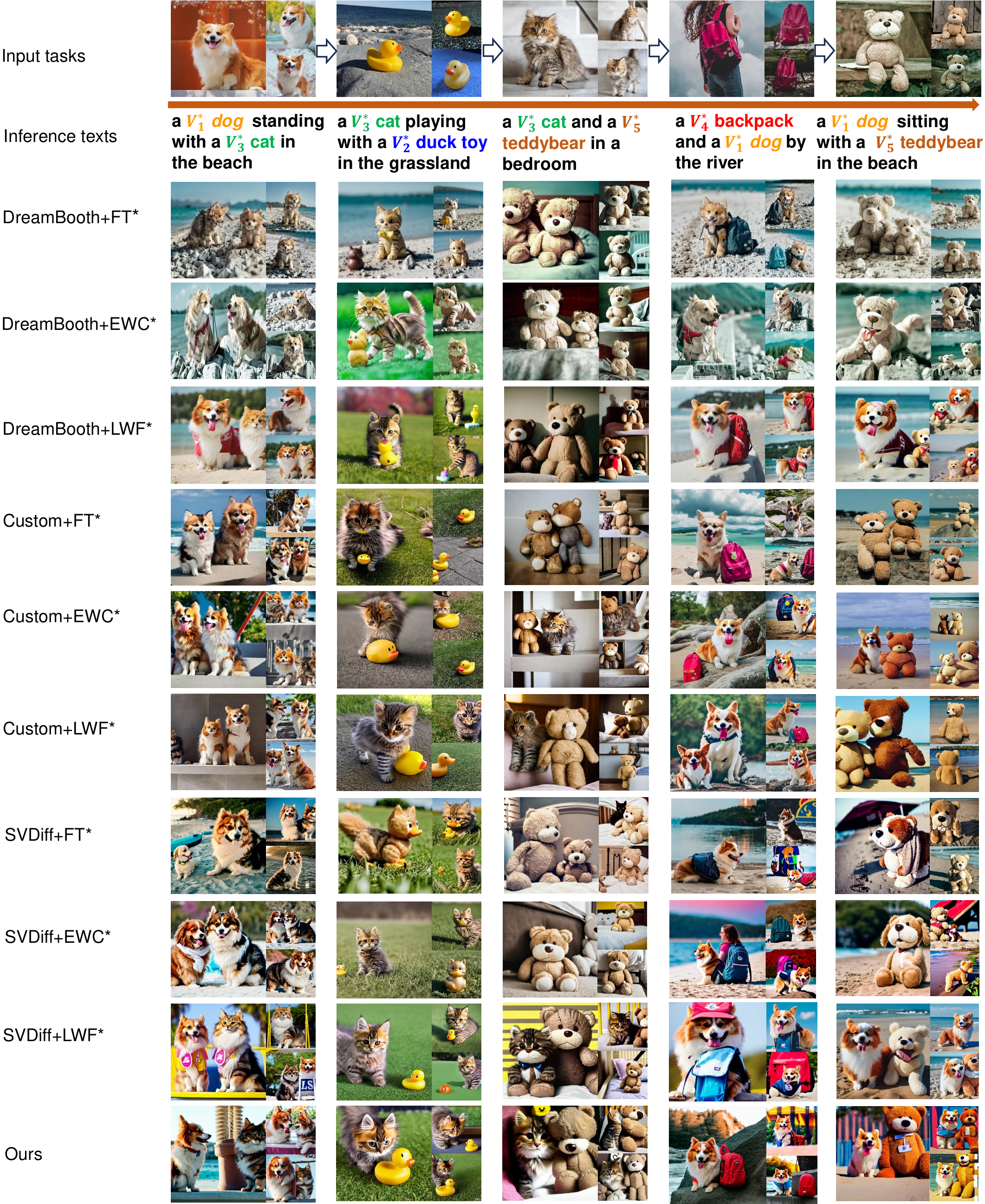}
        \caption{Qualitative comparisons between ours with the state-of-the-arts in the lifelong multi-concept generation setting, where the first two rows denote the continual user-specific generation tasks needed by the user, and the rest rows denotes the generation results among the competing methods for each same prompt.}
	\label{fig: experiments_multi}
	\vspace{-1pt}
\end{figure*}

%In contrast, Attend-and-Excite\cite{chefer2023attend} attempts to maximize the attention values for each object token, thereby effectively addressing the issue of catastrophic neglect. However, unrestricted amplification of the attention values can lead to catastrophic homogenization within L$^2$DM.

For the challenge posed by a significant representation disparity of two concepts, the proposed dynamic attend loss serves to harmonize and balance the representation of diverse concepts, which is defined as:
\begin{align}
\label{eq: eq8}
     \mc{L}_{\mr{DAL}} = \sum^{n_p}_{j=1}\big\{1-\mr{max}(\mc{G}(\frac{1}{n_l}\sum^{n_l}_{l=1}A_{t,l}^j))\big\},
\end{align}
where $n_p$ is the number of target personalized tokens (\emph{e.g.}, $V^*, V^\sim$). As defined in Eq.~\ref{eq: eq8}, we let all the target tokens participate to enhance the representation in an attention map, thereby further addressing the issue of concept-neglecting. Moreover, this loss ensures that the attention value assigned to any single concept does not become excessively large, thereby minimizing the risk of catastrophic concept-neglecting.

\textbf{Orthogonal Attention Artist (OAA):} Since the attribute-neglecting issue arises as a result of the mixing of attention distributions between personalized and prior concept tokens, we develop an orthogonal attention artist module to effectively disentangle the attention distributions of individual tokens. Drawing inspiration from LOSL\cite{patashnik2023localizing}, we explore the extraction of a mask from the attention distribution associated with one concept. This extracted mask is subsequently utilized to constrain the attention distribution pertaining to other concepts. In our work, we begin by carefully selecting the set $O_{t,l}^i$ by applying a fixed threshold, which signifies the high activation in the cross-attention score. Subsequently, we proceed to extract the concept-level mask in the following manner:
\begin{equation}
\begin{split}
	M^i_{t,l}[j,k] =
	\left\{
	\begin{aligned}		
		&1, \qquad \mr{if}~j\in O_{t,l}^i, \\
		&0, \qquad \mr{otherwise},  \\
	\end{aligned} 					
	\right.											
\end{split}							
\label{eq: eq9}	
\end{equation}
In particular, we extra the token-level mask of the prior concept token (\emph{e.g.,} dog) and the personalized token (\emph{e.g.,} V*sks) using Eq.~\ref{eq: eq9}. To further mitigate the potential mixing of attention distributions between personalized and object tokens, we employ the orthogonal attention loss. This artist module serves to incentivize the personalized token to solely represent in the location of the corresponding concept token. The orthogonal attention artist function can be simplified as:
\begin{align}
\label{eq: eq10}
     \mc{L}_{\mr{OAA}} =\frac{1}{n_ln_c}\sum^{n_l}_{l=1} \sum^{n_c}_{i=1}\sum^{n_p}_{j=1}
     \frac{\Vert \mathbb{I}_{\zeta}+ (-1)^{\mathbb{I}_{\zeta}}
     \cdot A^j_{t,l}  \cdot M_{t,l}^{i}\Vert^2_2} {\sum A^j_{t,l}},
\end{align}
where the binary function denoted $\mathbb{I}$ is defined such that it takes the value of 1 when the condition $i=j$ is satisfied, and 0 otherwise. As defined in Eq.~\ref{eq: eq10}, the orthogonal attention artist mechanism strengthens the representation of the personalized token when it matches the concept token (\emph{i.e.}, $i=j$), while constraining it when there is no match.

Overall, in the inference stage of our L$^2$DM, we introduce an artist objective that facilitates the progressive adjustment of $A^i_{t,l}$ and the subsequent updating of the latent code $z_t$ to ensure the independence and diversity of representation of target tokens. By leveraging the attention scores, we establish a guidance mechanism that enables the target token to effectively manifest within the latent code.

\section{Experiments} \label{sec: experiment}
To evaluate our lifelong text-to-image diffusion framework, we in this section present several comprehensive experiments with several state-of-the-arts, which encompass both lifelong single concept generation and lifelong multi-concept generation scenarios.

\subsection{Datasets and Evaluation}

\indent \textbf{Datasets:} We follow the dataset setting in DreamBooth~\cite{ruiz2023dreambooth} and Custom 
 Diffusion~\cite{kumari2022multi}, and adopt the collected 35 subjects in our experimental section, which contains unique pets and objects such as dogs, cats, backpacks and toys. For a fair comparison, we apply 20 prompts for each concept by following~\cite{kumari2022multi}. Moreover, we conduct our experiments on two lifelong generation task settings to evaluate the performance of our method. As shown in Fig.~\ref{fig: experiments_single}, we curate a five-tasks dataset by selecting five concepts from the built datasets in Custom Diffusion and DreamBooth.

%the second dataset

\begin{table*}[htbp]
\centering
\setlength{\tabcolsep}{1.6mm}
\setlength{\arraycolsep}{1pt}
\caption{Lifelong single-concept generation comparisons between ours with the state-of-the-arts in terms of Trainable Params, Text- and image-alignment ($\%$). Methods with the best and runner-up performance are marked as bolded red and blue color, respectively. }
%	\resizebox{\linewidth}{!}{
\scalebox{0.85}{
\begin{tabular}{l|c|ccccc|cl|ccccc|cl}
	\toprule
\makecell{\multirow{2}{*}{Comparison Methods}}  & \multirow{2}{*}{\#Params} & \multicolumn{7}{c|}{$\mr{IA}(\%)$} & \multicolumn{7}{c}{$\mr{TA}(\%)$} \\
	  & & dog & duck toy & cat & backpack & teddybear & Avg. & Imp.  & dog & duck toy & cat & backpack & teddybear & Avg. & Imp. \\
	\midrule
 Dreambooth\cite{ruiz2023dreambooth}+FT & 4.1M  &70.6 &70.0 &70.7 &74.7 &84.3  &74.1 &$\Uparrow$6.3 &23.1 &25.1 &21.2 &23.3 &26.5 &23.8 &$\Uparrow$6.3  \\
	Dreambooth\cite{ruiz2023dreambooth}+EWC\cite{li2017learning}  & 4.1M &79.1 &76.2 &71.8 &74.5 &80.2 &76.3&$\Uparrow$4.1 &24.3 &26.7 &25.8 &25.2 &26.9 &25.8 &$\Uparrow$4.3\\
	Dreambooth\cite{ruiz2023dreambooth}+LWF\cite{li2017learning}   & 4.1M &80.0 &76.8 &72.3 &76.7 &81.4 &77.4 &$\Uparrow$3.0 &25.7 &26.8 &27.4 &25.7 &27.5 &26.6 &$\Uparrow$3.5  \\
	% DDE \cite{Hu_2021_CVPR} &ViT-Base & 85.10M &   \\
	% AANets \cite{Liu2020AANets} &ViT-Base & 85.10M &    \\
     Custom\cite{kumari2022multi}+FT & 73.1M &76.5 &77.8 &70.3 &80.0 &84.5 &77.8 &$\Uparrow$2.6  &26.1 &26.3 &25.4 &25.5 &26.2 &25.9 &$\Uparrow$4.2 \\
 Custom\cite{kumari2022multi}+EWC\cite{li2017learning}  &73.1M &78.0 &77.4  &71.5 &80.4 &77.1 &78.1 &$\Uparrow$2.3 &26.1 &27.8 &22.3 &26.4 &27.9 &26.1 &$\Uparrow$4.0\\
 Custom\cite{kumari2022multi}+LWF\cite{li2017learning}  & 73.1M &79.0 &77.1 &71.4 &74.1 &78.7 &76.1 &$\Uparrow$4.3 &24.6 &28.5 &23.7 &25.8 &27.8 &26.1 &$\Uparrow$4.0 \\
	% DER \cite{Yan_2021_CVPR} (CVPR'2022) &ViT-Base & 85.10M & \\
        SVDiff\cite{SVDiff}+FT  & 1.7M &75.8 &76.4 &73.7 &80.0 &\textcolor[rgb]{1.00,0.00,0.00}{\textbf{85.7}}  &78.3 &$\Uparrow$2.1 &27.6 &26.8 &27.6 &26.7 &24.5 &26.6 &$\Uparrow$3.5 \\
	SVDiff\cite{SVDiff}+EWC\cite{li2017learning}  & 1.7M &78.4 &76.7 &72.3 &\textcolor[rgb]{1.00,0.00,0.00}{\textbf{82.8}}  &84.8 &79.0 &$\Uparrow$1.4 &28.9 &27.3 &29.5 &28.6 &27.2 &28.3 &$\Uparrow$1.8 \\
 	SVDiff\cite{SVDiff}+LWF\cite{li2017learning}  & 1.7M &83.4 &75.2 &74.8 &76.8 &85.5 &79.1 &$\Uparrow$1.3 &28.9 &28.9 &26.7 &29.4 &27.7 &28.3 &$\Uparrow$1.8 \\
        \midrule
        \rowcolor{lightgray}
        \textbf{Ours} & 1.7M  &\textcolor[rgb]{1.00,0.00,0.00}{\textbf{83.8}} &\textcolor[rgb]{1.00,0.00,0.00}{\textbf{78.0}} &\textcolor[rgb]{1.00,0.00,0.00}{\textbf{75.0}} &80.7 &84.5 &\textcolor[rgb]{1.00,0.00,0.00}{\textbf{80.4}} &\textbf{$\mathrm{-}$} & \textcolor[rgb]{1.00,0.00,0.00}{\textbf{30.6}} & \textcolor[rgb]{1.00,0.00,0.00}{\textbf{31.2}} &\textcolor[rgb]{1.00,0.00,0.00}{\textbf{30.0}} &\textcolor[rgb]{1.00,0.00,0.00}{\textbf{30.7}} &\textcolor[rgb]{1.00,0.00,0.00}{\textbf{28.0}} &\textcolor[rgb]{1.00,0.00,0.00}{\textbf{30.1}}&\textbf{$\mathrm{-}$} \\
        \midrule
\end{tabular}}
\label{tab: single_concept}
\end{table*}

\begin{table*}[htbp]
\centering
\setlength{\tabcolsep}{2.3mm}
\setlength{\arraycolsep}{1pt}
\caption{Lifelong multi-concept generation comparisons between ours with the state-of-the-arts in terms of Trainable Params, Text- and image-alignment ($\%$). Methods with the best and runner-up performance are marked as bolded red and blue color, respectively. The compared method with'*' denotes that this method performs multi-concept generation with the guidance of Attend-and-Excite~\cite{chefer2023attend}.}
%	\resizebox{\linewidth}{!}{
\scalebox{0.88}{
\begin{tabular}{l|c|ccccc|cl|ccccc|cl}
	\toprule
\makecell[c]{\multirow{2}{*}{Comparison Methods}}  & \multirow{2}{*}{\#Params} & \multicolumn{7}{c|}{$\mr{IA}(\%)$} & \multicolumn{7}{c}{$\mr{TA}(\%)$}  \\
	  & & D+C & C+Dt & C+T & B+D & D+T & Avg. & Imp. & D+C & C+Dt & C+T & B+D & D+T & Avg. & Imp. \\
	\midrule
Dreambooth\cite{ruiz2023dreambooth}+FT* & 4.1M  &60.2 &65.3 &66.3 &65.6 &69.1 &65.3 &$\Uparrow$9.8 &19.7 &18.3 &20.1 &19.5 &21.3 &19.8 &$\Uparrow$6.7  \\
	Dreambooth\cite{ruiz2023dreambooth}+EWC\cite{li2017learning}*  & 4.1M &62.8 &65.0 &65.2 &68.4 &70.5 &66.4 &$\Uparrow$8.7 &20.5 &20.5 &19.4 &21.2 &21.8 &20.7 &$\Uparrow$5.8\\
	Dreambooth\cite{ruiz2023dreambooth}+LWF\cite{li2017learning}*    & 4.1M &68.1 &71.8 &66.3 &70.7 &74.8 &70.3 &$\Uparrow$4.8 &23.9 &24.6 &23.5 &24.7 &25.6 &24.5 &$\Uparrow$2.0  \\

     Custom\cite{kumari2022multi}+FT* & 73.1M &66.7 &70.3 &68.0 &72.8 &75.0 &70.5 &$\Uparrow$4.6  &23.8 &26.2 &24.5 &23.1 &25.5 &24.6 &$\Uparrow$1.9  \\
 Custom\cite{kumari2022multi}+EWC\cite{li2017learning}*  &73.1M &67.8 &72.9 &68.9 &73.7 &73.4 &71.3 &$\Uparrow$3.8 &24.3 &24.7 &23.1 &23.5 &25.0 &24.1 &$\Uparrow$2.4\\
 Custom\cite{kumari2022multi}+LWF\cite{li2017learning}* & 73.1M &69.0 &71.8 &69.3 &74.1 &74.5 &71.7 &$\Uparrow$3.4 &23.8 &24.1 &25.0 &23.4 &27.2 &24.7 &$\Uparrow$1.8 \\

        SVDiff\cite{SVDiff}+FT*  & 1.7M &69.6 &70.4 &68.2 &76.3 &76.4 &72.2 &$\Uparrow$2.9 &24.6 &24.5 &23.2 &25.0 &25.1 &24.5 &$\Uparrow$2.0 \\
	SVDiff\cite{SVDiff}+EWC\cite{li2017learning}*  & 1.7M &68.7 &72.9 &71.4 &73.1 &77.7 &72.8 &$\Uparrow$2.3 &25.0 &24.2 &24.0 &24.8 &24.7 &24.5 &$\Uparrow$2.0 \\
 	SVDiff\cite{SVDiff}+LWF\cite{li2017learning}*  & 1.7M &68.7 &70.7 &67.0 &71.1 &75.5 &70.6 &$\Uparrow$4.5 &24.9 &26.5 &26.0 &26.8 &27.2 &25.4 &$\Uparrow$1.1 \\
        \midrule
        \rowcolor{lightgray}
        \textbf{Ours} & 1.7M  &\textcolor{deepred}{\textbf{70.1}} &\textcolor{deepred}{\textbf{75.0}} &\textcolor{deepred}{\textbf{73.5}} &\textcolor{deepred}{\textbf{76.9}} &\textcolor{deepred}{\textbf{80.1}} &\textcolor{deepred}{\textbf{75.1}} &\textbf{$\mathrm{-}$} &\textcolor{deepred}{\textbf{25.8}} &\textcolor{deepred}{\textbf{26.5}} &\textcolor{deepred}{\textbf{26.0}} &\textcolor{deepred}{\textbf{26.8}} &\textcolor{deepred}{\textbf{27.2}}&\textcolor{deepred}{\textbf{26.5}} &\textbf{$\mathrm{-}$} \\
        \midrule

\end{tabular}}
\label{tab: multi_concept}
\end{table*}

\begin{table}[t]
\setlength{\tabcolsep}{1.6mm}
\setlength{\arraycolsep}{1pt}
\caption{Lifelong single-concept generation comparisons between ours with the state-of-the-arts in terms of Trainable Params, TFR-IA($\%$) and TFR-TA($\%$). }
%	\resizebox{\linewidth}{!}{
\scalebox{0.84}{
\begin{tabular}{l|c|cl|cl}
	\toprule
\makecell{\multirow{1}{*}{Comparison Methods}}  & \multirow{1}{*}{\#Params}
	   & TFR-IA($\%$) & Imp.  & TFR-TA($\%$) & Imp. \\
	\midrule
 Dreambooth\cite{ruiz2023dreambooth}+FT & 4.1M  &8.3 &$\Downarrow$7.1 &2.8 &$\Downarrow$2.0   \\
	Dreambooth\cite{ruiz2023dreambooth}+EWC\cite{li2017learning}  & 4.1M &5.1 &$\Downarrow$3.9 &2.1 &$\Downarrow$1.3 \\
	Dreambooth\cite{ruiz2023dreambooth}+LWF\cite{li2017learning}   & 4.1M &3.2 &$\Downarrow$2.0 &2.2 &$\Downarrow$1.4   \\

     Custom\cite{kumari2022multi}+FT & 73.1M &4.3 &$\Downarrow$3.1 &2.4 &$\Downarrow$1.6   \\
 Custom\cite{kumari2022multi}+EWC\cite{li2017learning}  &73.1M &3.6 &$\Downarrow$2.4 &1.8 &$\Downarrow$1.0\\
 Custom\cite{kumari2022multi}+LWF\cite{li2017learning}  & 73.1M &4.6 &$\Downarrow$3.4 &2.1 &$\Downarrow$1.3  \\

        SVDiff\cite{SVDiff}+FT  & 1.7M &4.9 &$\Downarrow$3.7 &2.3 &$\Downarrow$1.5 \\
	SVDiff\cite{SVDiff}+EWC\cite{li2017learning}  & 1.7M &3.7 &$\Downarrow$3.5 &1.7 &$\Downarrow$0.9  \\
 	SVDiff\cite{SVDiff}+LWF\cite{li2017learning}  & 1.7M &3.0 &$\Downarrow$1.8 &1.4 &$\Downarrow$0.6  \\
        \midrule
        \rowcolor{lightgray}
        \textbf{Ours} & 1.7M  &\textcolor[rgb]{1.00,0.00,0.00}{\textbf{1.2}} &\textbf{$\mathrm{-}$}  &  \textcolor[rgb]{1.00,0.00,0.00}{\textbf{0.8}} &\textbf{$\mathrm{-}$} \\
        \midrule
\end{tabular}}
\label{tab: tfr}
\vspace{-10pt}
\end{table}

\indent\textbf{Evaluation metrics:} To fairly evaluate the generation performance in our lifelong learning setting, we generate four images per concept and per prompt for each generation task. After the lifelong diffusion model observes the last task, three metrics are adopted to evaluate the performance of lifelong generation: (1) \emph{Image-Alignment} (IA), following DreamBooth~\cite{ruiz2023dreambooth} and Custom Diffusion~\cite{kumari2022multi}, is the visual similarity computed in the feature space when we fed generated images with target concept and raw images into a pretrained CLIP-L model~\cite{DBLP:conf/icml/RadfordKHRGASAM21}; (2) \emph{Text-Alignment} (TA), can be obtained from CLIP-L feature space using the given text prompts and generated images; (3) \emph{Task Forgetting Rate} (TFR), can be computed as TFR-IA: $\mc{F}_{I}=\frac{1}{k-1}\sum_{\ell=1}^{k-1} \mc{I}_{\ell,\ell}-\mc{I}_{k,\ell}$ and TFR-TA: $\mc{F}_{T}=\frac{1}{k-1}\sum_{\ell=1}^{k-1} \mc{T}_{\ell,\ell}-\mc{T}_{k,\ell}$ by following \cite{diaz2018don}, where $\mc{I}_{\ell,\ell}$ and $\mc{T}_{\ell,\ell}$ denote IA and TA metrics of the $\ell$-th generation task, and $\mc{I}_{k,\ell}$ and $\mc{T}_{k,\ell}$ denote IA and TA metrics of the $\ell$-th task after learning the $k$-th generation task.

\subsection{Implementation Details}
We conduct lifelong generation experiments on our proposed L$^2$DM method and three representative baselines in personalized text-to-image generation, \emph{i.e.,} DreamBooth~\cite{ruiz2023dreambooth}, Custom Diffusion~\cite{kumari2022multi} (termed as Custom in this paper) and SVDiff~\cite{SVDiff}. Since there are currently no available methods that consider never-ending learning issues in text-to-image generation, we apply Fine-tuning (FT), LWF~\cite{li2017learning} and EWC~\cite{kirkpatrick2017overcoming} for each baseline in the lifelong learning setting. Both LWF~\cite{li2017learning} and EWC~\cite{kirkpatrick2017overcoming} are the representative lifelong learning methods. Meanwhile, it is unrealistic to employ the representative lifelong learning methods in DreamBooth~\cite{ruiz2023dreambooth} to achieve lifelong generation task, since catastrophic forgetting issue will destroy the whole diffusion model. We then apply LoRA\cite{hu2021lora}, as the compared method instead in this paper. Specifically, we train 300 steps for Custom~\cite{kumari2022multi} in each task, 500 steps for SVDiff~\cite{SVDiff} and Ours, and 800 steps for DreamBooth~\cite{ruiz2023dreambooth}. Futhermore, we apply a learning rate $1\times 10^{-5}$ for Custom, $1\times 10^{-4}$ for DreamBooth, $1\times 10^{-3}$ for SVDiff and $1.5\times 10^{-3}$ for Ours. Both DreamBooth and SVDiff methods are trained using a batch size of 1,  whereas Custom and Ours utilize a batch size of 2. The text encoders of all methods are trained during the optimization process, where Custom method expands additional token as a personalized token (\emph{i.e.,} $\mr{V}^*$), and other methods use a rarely occurring token (\emph{e.g.}, sks). As for the prior preservation loss in Eq.~\ref{eq: pdm}, 200 images for each new concept are generated by Stable Diffusion. To reduce computational cost, we randomly choose 50 images for each learned concept to compute $\mc{L}^{k}_{\mr{PR}}$ in Eq.~\ref{eq: eq3}. As for the inference stage, we apply a DDPM sampler with 200 steps and a classifier-free guidance scale of 7 for all the methods. To be fair, we employ the sate-of-the-art method Attend-and-Excite~\cite{chefer2023attend} to all competing methods to generate multi-concept images, \emph{e.g.}, SVDiff+FT*.

\begin{figure*}[t]
	\centering
	\includegraphics[width=520pt, height=130pt]
	{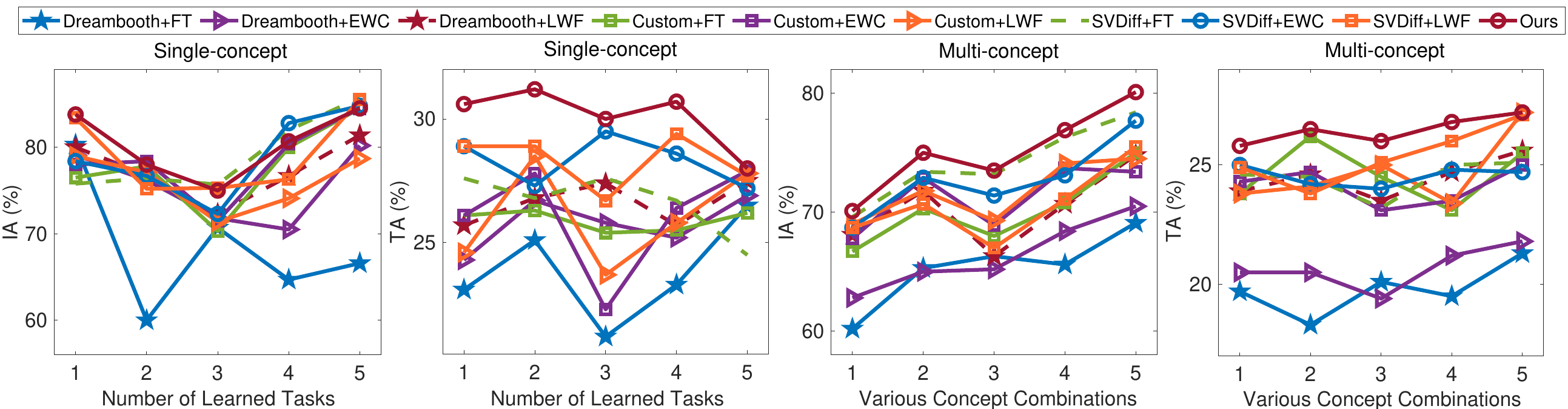}
        \caption{Text-and image-alignment for lifelong single-concept (left) and multi-concept (right) generation setting. Compared with the baseline methods (\emph{e.g., } Dreambooth\cite{ruiz2023dreambooth}+EWC\cite{li2017learning}, Custom\cite{kumari2022multi}+EWC\cite{li2017learning}, SVDiff\cite{SVDiff}+LWF\cite{li2017learning}), our L$^2$DM model is better than the baselines in terms of both IA($\%$) and TA($\%$). }
	\label{fig: ia_ta}
\end{figure*}

\begin{figure*}[t]
	\centering
	\includegraphics[width=520pt, height=130pt]
	{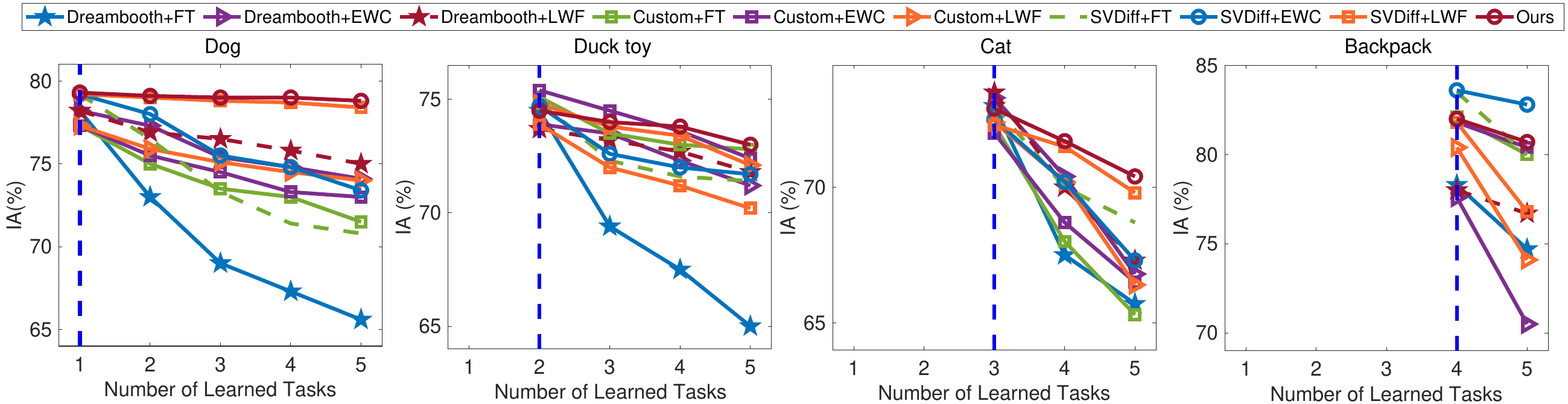}
        \caption{Task forgetting rate comparisons for lifelong single-concept generation setting. Compared with the baseline methods, our L$^2$DM model can achieve the best performance in most cases, especially on the earlier learned tasks.}
	\label{fig: forgetting}
 \vspace{-10pt}
\end{figure*}

\subsection{Comparison Evaluation}
We in this subsection test the competing models in the lifelong single-concept and lifelong multi-concept generation scenarios with a set of challenging prompts, followed by the ablation studies of our proposed L$^2$DM model.

\subsubsection{Lifelong Single-concept Generation Results}\label{sec:single-concept}
To evaluate the lifelong learning property in the text-to-image diffusion issue, we define the task sequence of learning and synthesizing each new concept in lifelong learning manner as follows: (1) dog, (2) duck toy, (3) cat, (4) backpack, (5) teddybear. After learning all the personalized concepts, we then demonstrate the results of generating each new target concepts on the same prompt. We show the superior results of our lifelong text-to-image diffusion model in respect of \textbf{Qualitative Comparisons} and \textbf{Quantitative Comparisons}.

\textbf{Qualitative Comparisons:} As the images shown in Fig.~\ref{fig: experiments_single}, our proposed could capture the visual details for the target concepts while achieving better text-image alignment. This observation indicates the effectiveness of concept-specific memory and concept-specific distillation modules in consolidating previous diffusion knowledge. On the contrary, most competing methods seriously suffers from the great catastrophic forgetting on the prior concepts or past encountered concepts, \emph{e.g.,}, DreamBooth~\cite{ruiz2023dreambooth}+FT, Custom~\cite{kumari2022multi}+FT and SVDiff~\cite{SVDiff}+FT. The main reason is that simply adopting fine-tuning strategy with the existing personalized text-to-image diffusion models fails to remember or access the previous knowledge, and cannot merge all the specific concepts in the end. In the case of the first column, we find forgetting of the first concept except for our model, DreamBooth+LWF and SVDiff+LWF, which illustrates that almost all the methods overfit to the new generation tasks, even combing with the classical lifelong learning methods.
%overfit to new task and result in suffering from  when directly applied to lifelong generation task . Moreover, almost all methods cannot recall  what the first learned dog looks like after observe all tasks except Dreambooth+LWF, SVDiff+LWF and Ours.

\begin{figure*}[htbp]
	\centering
	\includegraphics[width=515pt, height=320pt]
	{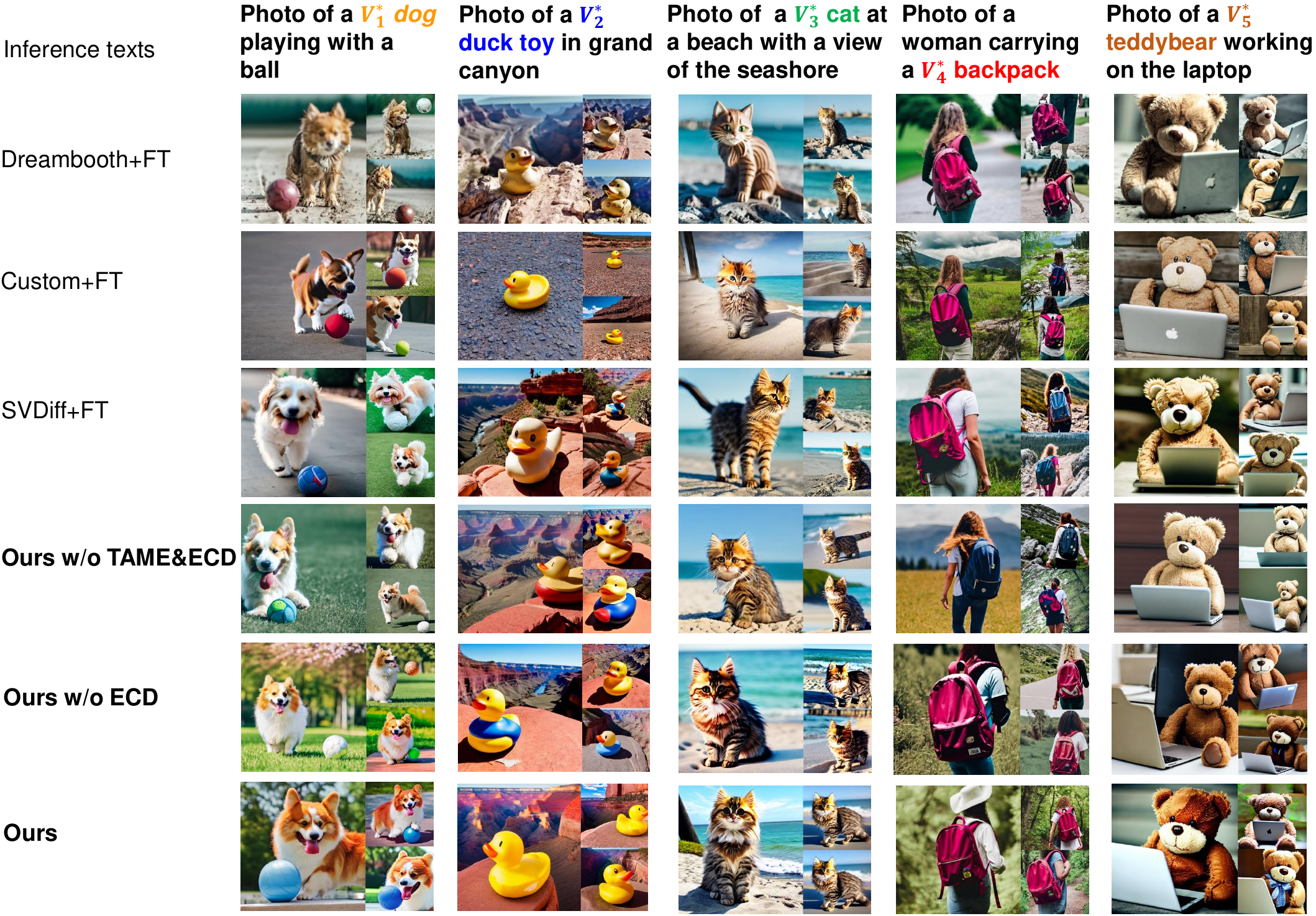}
        \caption{Ablation studies of our catastrophic forgetting regularization, where TAME and ECD denotes the Task-Aware Memory Enhancement and Elastic  Concept Distillation modules, respectively.}
	\label{fig: ablation_1}
	\vspace{-10pt}
\end{figure*}

\begin{figure*}[htbp]
	\centering
	\includegraphics[width=515pt, height=320pt]
	{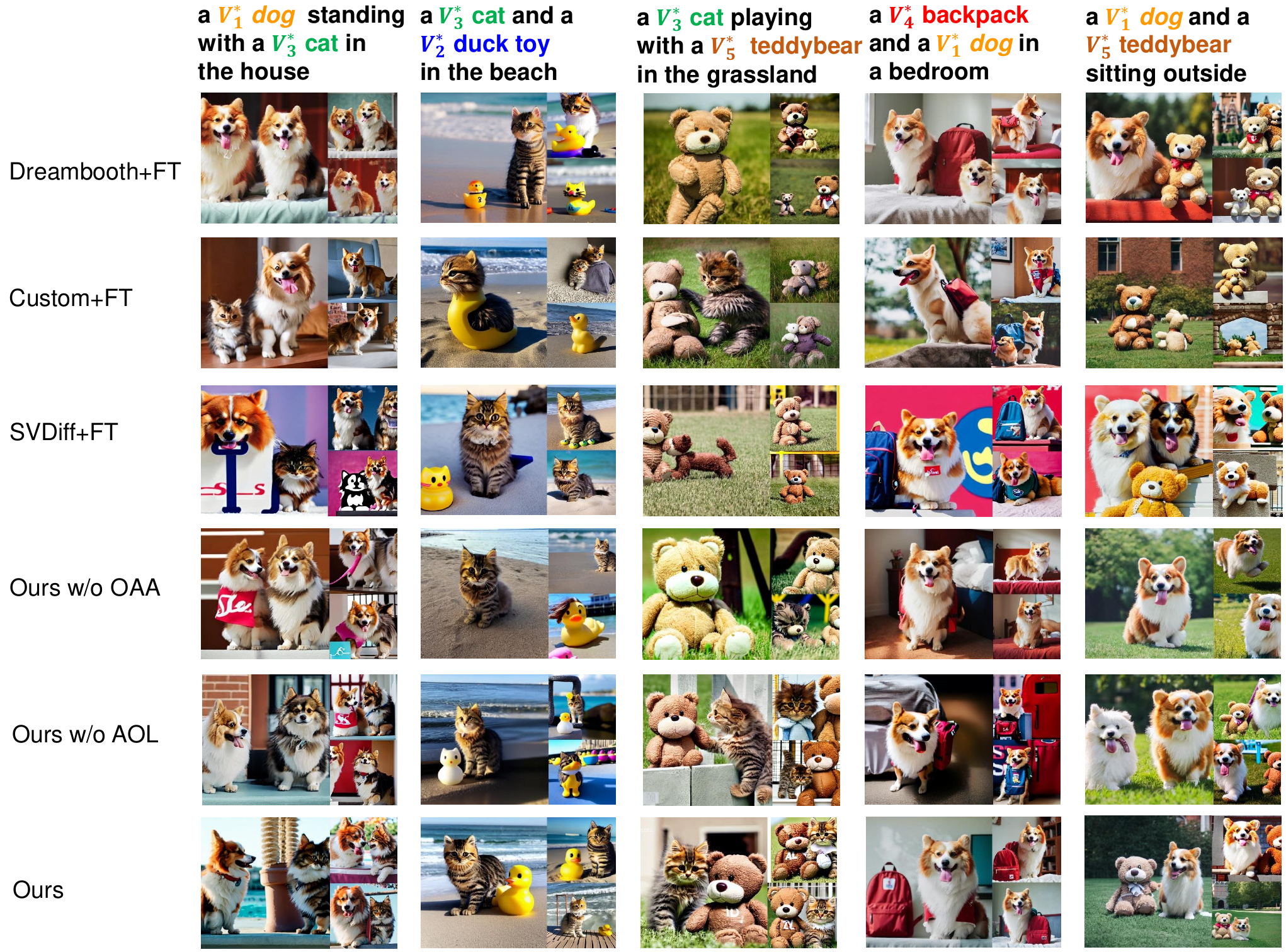}
        \caption{Ablation studies of our catastrophic neglecting regularization, where CAA and OAA denotes the concept attention artist and orthogonal attention artist modules, respectively.}
	\label{fig: ablation_2}
	\vspace{-10pt}
\end{figure*}

\begin{figure*}[htbp]
	\centering

	\includegraphics[width=500pt]
	{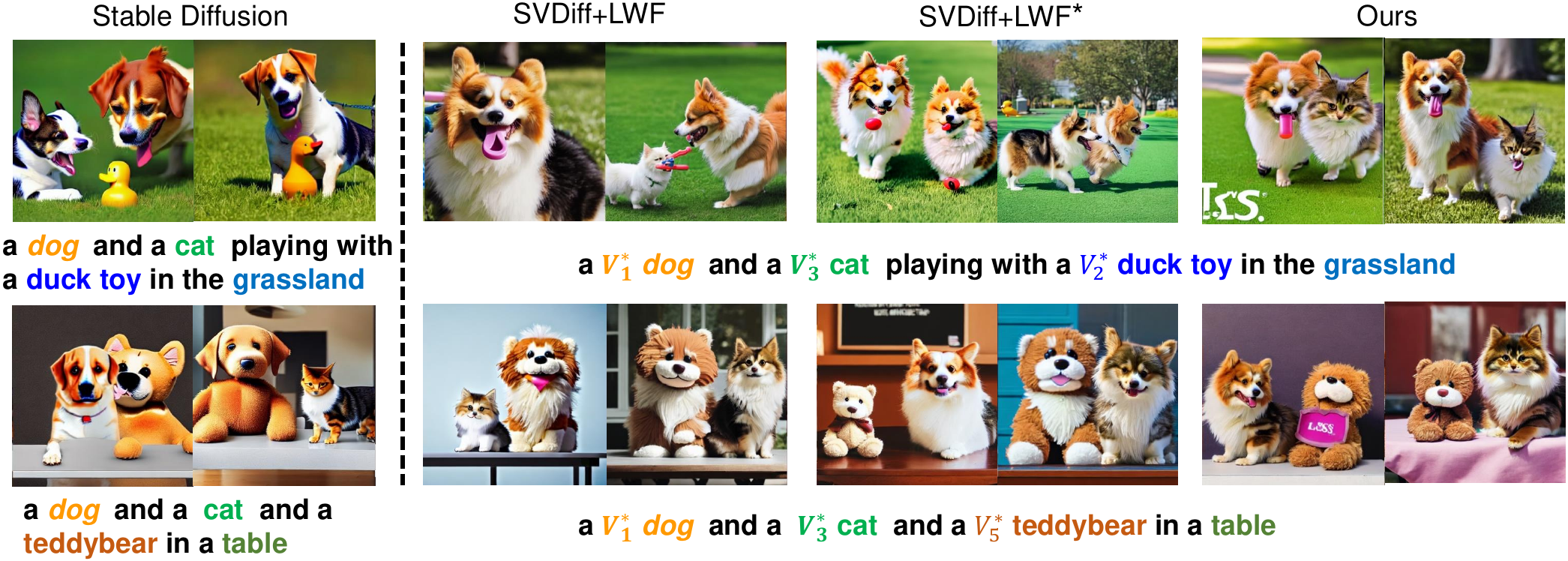}
  \vspace{-5pt}
        \caption{Failure cases in the lifelong text-to-image diffusion setting, where Stable Diffusion model generates the corresponding user-specific concepts, and the rest methods with a same text prompt adopts the concept images provided by user.}
	\label{fig: limitation}

 	\vspace{-10pt}
\end{figure*}
\textbf{Quantitative Comparisons:} To quantitatively evaluate the performance of each competing model, we generate 500 images on 10 text prompts by following Custom Diffusion~\cite{kumari2022multi} model, \emph{i.e.,} 50 samples per prompt. After setting the task sequence of learning each new concept as: (1) dog, (2) duck toy, (3) cat, (4) backpack, (5) teddybear, we evaluate all the competing methods in terms of IA ($\%$) and TA ($\%$) for the image-alignment and text-alignment performance, respectively. As the results shown in Table.~\ref{tab: single_concept} and Fig.~\ref{fig: ia_ta}, we have the following conclusions: 1) the average performance of our model outperforms all compared methods by $1.3\%~6.3\%$ IA and $1.8\%~6.3\%$ TA on various generation tasks. Such improvements suggest less overfitting of our model to the new concept, and justifies the effectiveness of utilizing concept-specific distillation to reduce catastrophic forgetting; 2) the updated network parameters is only 1.7MB in our model, which is lowest network parameters in comparison to other models. This metric means our model needs more lower requirements for learning new concepts consecutively; 3) we also present the task forgetting rates in Fig.~\ref{fig: forgetting} and Table.~\eqref{tab: tfr} after calculating $\mc{F}_{I}$ and $\mc{F}_{T}$ , which could reflect the performance of each concept after learning different generation tasks. Notice that our model  attains the strongest anti-forgetting ability in the lifelong text-to-image diffusion setting, which further lends the effectiveness of our proposed model in storing diffusion knowledge.

%by generating 500 images on 10 text prompts, where each task tries to generate a new single-concept. The task sequence of learning each new concept in lifelong learning manner is in a same setting: (1) dog, (2) duck toy, (3) cat, (4) backpack, (5) teddybear. After learning a new concept given by user, we obtain 500 images using the same prompt for each learned concept to compute $\mc{F}_{k}$ and $\mc{F}_{k}$. As published in Table.~\ref{tab: single_concept} and Fig.~\ref{fig: ia_ta}, the average results of ours outperform all compared methods in terms of IA and TA. What's more, the proposed method achieves a significant improvement in old concepts, which represents our method can effectively reduce catastrophic forgetting. We present more results in Fig.~\ref{fig: forgetting}, which reflects the performance of each concept after learning different tasks. We can observe that the curve of ours has a lower rate of decline than other baselines.
%What's more, the proposed method achieves a significant improvement in old concepts, which represents our method can effectively reduce catastrophic forgetting. Moreover, we present more results in Fig.~\ref{fig: forgetting}, which reflects the performance of each concept after learning different tasks. We can observe that the curve of ours has a lower rate of decline than other baselines.

\subsubsection{Lifelong Multi-concept Generation Results}\label{sec:multi-concept}
To evaluate the lifelong learning property in generating multi-concept in the same scene, we also define the task sequence of learning new concept as: (1) dog, (2) duck toy, (3) cat, (4) backpack, (5) teddybear. The multi-concept combinations in this paper consists of Dog+Cat(D+C), {Cat+Duck toy(C+Dt)}, {Cat+Teddybear(C+T)}, {Backpack+Dog(B+D)} and {Dog+Teddybear(D+T)}. To be specific, we compare ours with baselines using a generated set consisting of 400 images with 8 prompts, and present these in Fig.~\ref{fig: experiments_multi} after fairly selecting some representative images. From the Fig.~\ref{fig: experiments_multi}, our proposed model outperforms all methods with the same prompt, which shows the importance of object attention artist for multi-concept generation. In the setting of lifelong generation tasks, we observe the catastrophic forgetting and catastrophic neglect amongst most of baselines, since these methods are difficult to backtrack the personalized concepts and prevent the influence of one concept on the other concepts. This also results in generating some wrong images, \emph{i.e.,} the low text-and image-alignment. To further show the text-and image-alignment more intuitive, we provide the corresponding results in Table.~\ref{tab: multi_concept} and Fig.~\ref{fig: ia_ta}. Notice that our proposed model achieves a $2.7\%\sim 16.4\%$ IA and $0.7\%\sim 9.5\%$ TA, compared to all the baselines. The results above directly reflect that the proposed methods achieve a better performance in lifelong multi-concept generation, which is consistent with the results in the lifelong single-concept generation setting.

%We compare ours with baselines using a generated set consisting of 400 images with 8 prompts. The multi-concepts pairs consists of 'Dog+Cat', 'Cat+Duck toy', 'Cat+Teddybear', 'Backpack+Dog' and 'Dog+Teddybear'. We fairly select some representative images and present these in Fig.~\ref{fig: experiments_multi}. Obviously, suffering from catastrophic forgetting, neglect and homogenization, most of baselines are difficult to recall the personalized concepts and prevent the influence of one concept on the other concept, resulting in generating some wrong images. In order to show it more intuitively, we give some results in Table.~\ref{tab: multi_concept} and Fig.~\ref{fig: ia_ta}. The results and covers directly reflect that the proposed methods achieve a better performance in lifelong and multi-concepts generation than baselines.

\subsubsection{Ablation Study}
We in this section present the ablation studies of our proposed method to show the contribution of each component. Each ablation study comparisons are justified with the same setup as Sec.~\ref{sec:single-concept} and Sec.~\ref{sec:multi-concept}.

\textbf{Without Catastrophic Forgetting Regularization:} As detailed in Sec.~\ref{subsec:our_framework}, we propose two modules to tackle catastrophic forgetting on encountered concepts (\emph{i.e.}, Task-Aware Memory Enhancement and Elastic Concept Distillation modules). Here, we eliminate them one by one to evaluate the effect of each module, and compare \emph{Ours}, \emph{Ours w/o TAME}, \emph{Ours w/o ECD} with baselines based on fine-tuning. As the results shown in Fig.~\ref{fig: ablation_1}, \emph{Ours w/o TAME$\&$ECD} has a lower image-alignment for the earlier concepts, and tends to forget the details of encountered concepts in comparison to Ours model. However, \emph{Ours w/o ECD} performs better than \emph{Ours w/o TAME$\&$ECD} in the color details for target concept, \emph{e.g.,} the first concept [$V_1^*$ dog]. These results above lends that both TAME and ECD play a key role in mitigating catastrophic forgetting, which verifies the rationality of Ours.

\textbf{Without Catastrophic Neglecting Regularization:} To comprehensively evaluate the performance of our method with this setup, \emph{i.e.}, using multi-concept to generate images, we compare \emph{Ours}, \emph{Ours w/o CAA}, \emph{Ours w/o OAA} with DreamBooth+LWF, Custom+LWF, SVDiff+LWF. From the present results in Fig.~\ref{fig: ablation_2}, we can notice that Our model can well address both catastrophic forgetting and catastrophic neglecting issues than baselines. The \emph{Ours w/o CAA} model lead to a lower text-alignment with the text prompt, and further cause one concept of the prompt is not generated,\emph{e.g.,} ``a [$V_1^*$ dog] and a [$V_5^*$ teddybear] sitting outside''. Although the \emph{Ours w/o OAA} model can generate correct concept number on the text prompt, it binds attributes or visual details on the wrong concepts. Different from the methods above, our model can generate a higher text-and image-alignment in the qualitative view. This also justifies that our proposed CAA and OAA modules play a key role in synthesizing multi-concepts images.

% \subsubsection{Limitations}

% As shown in Fig~\ref{fig: limitation}, we present some failure generated images using complicated prompts consisting of $3\sim4$ concepts. All compared methods and ours suffer  catastrophic neglecting when composing four or more concepts together. Furthermore, original Stable Diffusion also faces the same limitations.
 \section{Conclusion}\label{sec:conclusion}
We in this paper explores how to continually learn new user-specific concepts and perform lifelong text-to-image generation tasks with a pre-trained diffusion model. Our proposed L$^2$DM model can efficiently attain consecutive concept learning tasks without damaging existing generation system, which is different from existing personalized text-to-image diffusion model. To be specific, our L$^2$DM model is developed via considering two aspects, \emph{i.e.,} catastrophic forgetting and catastrophic neglecting. The catastrophic forgetting issue can be well mitigated with our proposed task-ware memory enhancement and elastic concept distillation modules; the catastrophic neglecting issue in the multi-concept generation stage is explored and further addressed with a concept attention artist and an orthogonal attention artist module. Comprehensive experiments are conducted to show that our model can achieves a significant performance in terms of text-alignment, image-alignment and task forgetting rate. 

\textbf{Limitations:} As shown in Fig.~\ref{fig: limitation}, some difficult compositions remain challenging and are synthesized wrongly, which is similar with the original Stable Diffusion model. In addition, all the compared methods and ours suffer from heavy catastrophic neglecting issue when composing four or more concepts together. This neglect issue will be our next research direction in the future. 

\ifCLASSOPTIONcompsoc
 % \section*{Acknowledgments}
\else
\fi

%The authors would like to thank Prof. Eric Eaton for his constructive suggestion. The research was supported by National Natural Science Foundation of China under Grant (61722311, U1613214, 61821005, 61533015) and LiaoNing Revitalization Talents Program (XLYC1807053).

%This work is supported by National Natural Science Foundation of China under Grant (61722311, U1613214, 61821005, 61533015) and LiaoNing Revitalization Talents Program (XLYC1807053)
\ifCLASSOPTIONcaptionsoff
  \newpage
\fi

\begin{comment}

\end{comment}

\vspace{-0.01pt}
\bibliographystyle{IEEEtran}
\bibliography{StableDiffusion,diffusion,Multitask}

\begin{thebibliography}{10}
\providecommand{\url}[1]{#1}
\csname url@samestyle\endcsname
\providecommand{\newblock}{\relax}
\providecommand{\bibinfo}[2]{#2}
\providecommand{\BIBentrySTDinterwordspacing}{\spaceskip=0pt\relax}
\providecommand{\BIBentryALTinterwordstretchfactor}{4}
\providecommand{\BIBentryALTinterwordspacing}{\spaceskip=\fontdimen2\font plus
\BIBentryALTinterwordstretchfactor\fontdimen3\font minus \fontdimen4\font\relax}
\providecommand{\BIBforeignlanguage}[2]{{%
\expandafter\ifx\csname l@#1\endcsname\relax
\typeout{** WARNING: IEEEtran.bst: No hyphenation pattern has been}%
\typeout{** loaded for the language `#1'. Using the pattern for}%
\typeout{** the default language instead.}%
\else
\language=\csname l@#1\endcsname
\fi
#2}}
\providecommand{\BIBdecl}{\relax}
\BIBdecl

\bibitem{xu2023open}
J.~Xu, S.~Liu, A.~Vahdat, W.~Byeon, X.~Wang, and S.~De~Mello, ``Open-vocabulary panoptic segmentation with text-to-image diffusion models,'' in \emph{Proceedings of the IEEE/CVF Conference on Computer Vision and Pattern Recognition}, 2023, pp. 2955--2966.

\bibitem{blattmann2023align}
A.~Blattmann, R.~Rombach, H.~Ling, T.~Dockhorn, S.~W. Kim, S.~Fidler, and K.~Kreis, ``Align your latents: High-resolution video synthesis with latent diffusion models,'' in \emph{Proceedings of the IEEE/CVF Conference on Computer Vision and Pattern Recognition}, 2023, pp. 22\,563--22\,575.

\bibitem{gu2022vector}
S.~Gu, D.~Chen, J.~Bao, F.~Wen, B.~Zhang, D.~Chen, L.~Yuan, and B.~Guo, ``Vector quantized diffusion model for text-to-image synthesis,'' in \emph{CVPR}, 2022, pp. 10\,696--10\,706.

\bibitem{balaji2022ediffi}
Y.~Balaji, S.~Nah, X.~Huang, A.~Vahdat, J.~Song, K.~Kreis, M.~Aittala, T.~Aila, S.~Laine, B.~Catanzaro \emph{et~al.}, ``ediffi: Text-to-image diffusion models with an ensemble of expert denoisers,'' \emph{arXiv preprint arXiv:2211.01324}, 2022.

\bibitem{rombach2022high}
R.~Rombach, A.~Blattmann, D.~Lorenz, P.~Esser, and B.~Ommer, ``High-resolution image synthesis with latent diffusion models,'' in \emph{Proceedings of the IEEE/CVF Conference on Computer Vision and Pattern Recognition}, 2022, pp. 10\,684--10\,695.

\bibitem{wang2022diffusiondb}
Z.~J. Wang, E.~Montoya, D.~Munechika, H.~Yang, B.~Hoover, and D.~H. Chau, ``Diffusiondb: A large-scale prompt gallery dataset for text-to-image generative models,'' \emph{arXiv preprint arXiv:2210.14896}, 2022.

\bibitem{nichol2021glide}
A.~Nichol, P.~Dhariwal, A.~Ramesh, P.~Shyam, P.~Mishkin, B.~McGrew, I.~Sutskever, and M.~Chen, ``Glide: Towards photorealistic image generation and editing with text-guided diffusion models,'' \emph{arXiv preprint arXiv:2112.10741}, 2021.

\bibitem{saharia2022photorealistic}
C.~Saharia, W.~Chan, S.~Saxena, L.~Li, J.~Whang, E.~L. Denton, K.~Ghasemipour, R.~Gontijo~Lopes, B.~Karagol~Ayan, T.~Salimans \emph{et~al.}, ``Photorealistic text-to-image diffusion models with deep language understanding,'' \emph{Advances in Neural Information Processing Systems}, vol.~35, pp. 36\,479--36\,494, 2022.

\bibitem{thrun1996discovering}
S.~Thrun and J.~O'Sullivan, ``Discovering structure in multiple learning tasks: The tc algorithm,'' in \emph{ICML}, 1996, pp. 489--497.

\bibitem{thrun2012explanation}
S.~Thrun, \emph{Explanation-based neural network learning: A lifelong learning approach}.\hskip 1em plus 0.5em minus 0.4em\relax Springer Science \& Business Media, 2012, vol. 357.

\bibitem{Sun9037204}
G.~{Sun}, Y.~{Cong}, Y.~{Zhang}, G.~{Zhao}, and Y.~{Fu}, ``Continual multiview task learning via deep matrix factorization,'' \emph{IEEE Transactions on Neural Networks and Learning Systems}, pp. 1--12, 2020.

\bibitem{ammar2015autonomous}
H.~B. Ammar, E.~Eaton, J.~M. Luna, and P.~Ruvolo, ``Autonomous cross-domain knowledge transfer in lifelong policy gradient reinforcement learning,'' in \emph{International Joint Conference on Artificial Intelligence}, 2015, pp. 3345--3351.

\bibitem{kirkpatrick2017overcoming}
J.~Kirkpatrick, R.~Pascanu, N.~Rabinowitz, J.~Veness, G.~Desjardins, A.~A. Rusu, K.~Milan, J.~Quan, T.~Ramalho, A.~Grabska-Barwinska \emph{et~al.}, ``Overcoming catastrophic forgetting in neural networks,'' \emph{Proceedings of the national academy of sciences}, vol. 114, no.~13, pp. 3521--3526, 2017.

\bibitem{li2016learning}
Z.~Li and D.~Hoiem, ``Learning without forgetting,'' in \emph{European Conference on Computer Vision}, 2016, pp. 614--629.

\bibitem{yoon2018lifelong}
J.~Yoon, E.~Yang, J.~Lee, and S.~J. Hwang, ``Lifelong learning with dynamically expandable networks,'' in \emph{Sixth International Conference on Learning Representations}.\hskip 1em plus 0.5em minus 0.4em\relax ICLR, 2018.

\bibitem{yan2021dynamically}
S.~Yan, J.~Xie, and X.~He, ``Der: Dynamically expandable representation for class incremental learning,'' in \emph{Proceedings of the IEEE/CVF Conference on Computer Vision and Pattern Recognition}, 2021, pp. 3014--3023.

\bibitem{douillard2022dytox}
A.~Douillard, A.~Ram{\'e}, G.~Couairon, and M.~Cord, ``Dytox: Transformers for continual learning with dynamic token expansion,'' in \emph{Proceedings of the IEEE/CVF Conference on Computer Vision and Pattern Recognition}, 2022, pp. 9285--9295.

\bibitem{song2020denoising}
J.~Song, C.~Meng, and S.~Ermon, ``Denoising diffusion implicit models,'' \emph{arXiv preprint arXiv:2010.02502}, 2020.

\bibitem{nichol2021improved}
A.~Q. Nichol and P.~Dhariwal, ``Improved denoising diffusion probabilistic models,'' in \emph{International Conference on Machine Learning}.\hskip 1em plus 0.5em minus 0.4em\relax PMLR, 2021, pp. 8162--8171.

\bibitem{saharia2022palette}
C.~Saharia, W.~Chan, H.~Chang, C.~Lee, J.~Ho, T.~Salimans, D.~Fleet, and M.~Norouzi, ``Palette: Image-to-image diffusion models,'' in \emph{ACM SIGGRAPH 2022 Conference Proceedings}, 2022, pp. 1--10.

\bibitem{wang2022pretraining}
T.~Wang, T.~Zhang, B.~Zhang, H.~Ouyang, D.~Chen, Q.~Chen, and F.~Wen, ``Pretraining is all you need for image-to-image translation,'' \emph{arXiv preprint arXiv:2205.12952}, 2022.

\bibitem{singer2022make}
U.~Singer, A.~Polyak, T.~Hayes, X.~Yin, J.~An, S.~Zhang, Q.~Hu, H.~Yang, O.~Ashual, O.~Gafni \emph{et~al.}, ``Make-a-video: Text-to-video generation without text-video data,'' \emph{arXiv preprint arXiv:2209.14792}, 2022.

\bibitem{poole2022dreamfusion}
B.~Poole, A.~Jain, J.~T. Barron, and B.~Mildenhall, ``Dreamfusion: Text-to-3d using 2d diffusion,'' \emph{arXiv preprint arXiv:2209.14988}, 2022.

\bibitem{lin2023magic3d}
C.-H. Lin, J.~Gao, L.~Tang, T.~Takikawa, X.~Zeng, X.~Huang, K.~Kreis, S.~Fidler, M.-Y. Liu, and T.-Y. Lin, ``Magic3d: High-resolution text-to-3d content creation,'' in \emph{Proceedings of the IEEE/CVF Conference on Computer Vision and Pattern Recognition}, 2023, pp. 300--309.

\bibitem{radford2021learning}
A.~Radford, J.~W. Kim, C.~Hallacy, A.~Ramesh, G.~Goh, S.~Agarwal, G.~Sastry, A.~Askell, P.~Mishkin, J.~Clark \emph{et~al.}, ``Learning transferable visual models from natural language supervision,'' in \emph{International conference on machine learning}.\hskip 1em plus 0.5em minus 0.4em\relax PMLR, 2021, pp. 8748--8763.

\bibitem{esser2021taming}
P.~Esser, R.~Rombach, and B.~Ommer, ``Taming transformers for high-resolution image synthesis,'' in \emph{Proceedings of the IEEE/CVF conference on computer vision and pattern recognition}, 2021, pp. 12\,873--12\,883.

\bibitem{feng2022training}
W.~Feng, X.~He, T.-J. Fu, V.~Jampani, A.~Akula, P.~Narayana, S.~Basu, X.~E. Wang, and W.~Y. Wang, ``Training-free structured diffusion guidance for compositional text-to-image synthesis,'' \emph{arXiv preprint arXiv:2212.05032}, 2022.

\bibitem{chefer2023attend}
H.~Chefer, Y.~Alaluf, Y.~Vinker, L.~Wolf, and D.~Cohen-Or, ``Attend-and-excite: Attention-based semantic guidance for text-to-image diffusion models,'' \emph{arXiv preprint arXiv:2301.13826}, 2023.

\bibitem{ruiz2023dreambooth}
N.~Ruiz, Y.~Li, V.~Jampani, Y.~Pritch, M.~Rubinstein, and K.~Aberman, ``Dreambooth: Fine tuning text-to-image diffusion models for subject-driven generation,'' in \emph{Proceedings of the IEEE/CVF Conference on Computer Vision and Pattern Recognition}, 2023, pp. 22\,500--22\,510.

\bibitem{gal2022image}
R.~Gal, Y.~Alaluf, Y.~Atzmon, O.~Patashnik, A.~H. Bermano, G.~Chechik, and D.~Cohen-Or, ``An image is worth one word: Personalizing text-to-image generation using textual inversion,'' \emph{arXiv preprint arXiv:2208.01618}, 2022.

\bibitem{kumari2022multi}
N.~Kumari, B.~Zhang, R.~Zhang, E.~Shechtman, and J.-Y. Zhu, ``Multi-concept customization of text-to-image diffusion,'' \emph{arXiv preprint arXiv:2212.04488}, 2022.

\bibitem{SVDiff}
\BIBentryALTinterwordspacing
L.~Han, Y.~Li, H.~Zhang, P.~Milanfar, D.~N. Metaxas, and F.~Yang, ``Svdiff: Compact parameter space for diffusion fine-tuning,'' \emph{CoRR}, vol. abs/2303.11305, 2023. [Online]. Available: \url{https://doi.org/10.48550/arXiv.2303.11305}
\BIBentrySTDinterwordspacing

\bibitem{ramesh2022hierarchical}
A.~Ramesh, P.~Dhariwal, A.~Nichol, C.~Chu, and M.~Chen, ``Hierarchical text-conditional image generation with clip latents,'' \emph{arXiv preprint arXiv:2204.06125}, 2022.

\bibitem{li2017learning}
Z.~Li and D.~Hoiem, ``Learning without forgetting,'' \emph{IEEE transactions on pattern analysis and machine intelligence}, vol.~40, no.~12, pp. 2935--2947, 2017.

\bibitem{meng2023distillation}
C.~Meng, R.~Rombach, R.~Gao, D.~Kingma, S.~Ermon, J.~Ho, and T.~Salimans, ``On distillation of guided diffusion models,'' in \emph{Proceedings of the IEEE/CVF Conference on Computer Vision and Pattern Recognition}, 2023, pp. 14\,297--14\,306.

\bibitem{patashnik2023localizing}
O.~Patashnik, D.~Garibi, I.~Azuri, H.~Averbuch-Elor, and D.~Cohen-Or, ``Localizing object-level shape variations with text-to-image diffusion models,'' \emph{arXiv preprint arXiv:2303.11306}, 2023.

\bibitem{DBLP:conf/icml/RadfordKHRGASAM21}
\BIBentryALTinterwordspacing
A.~Radford, J.~W. Kim, C.~Hallacy, A.~Ramesh, G.~Goh, S.~Agarwal, G.~Sastry, A.~Askell, P.~Mishkin, J.~Clark, G.~Krueger, and I.~Sutskever, ``Learning transferable visual models from natural language supervision,'' in \emph{Proceedings of the ICML}, ser. Proceedings of Machine Learning Research, M.~Meila and T.~Zhang, Eds., vol. 139.\hskip 1em plus 0.5em minus 0.4em\relax {PMLR}, 2021, pp. 8748--8763. [Online]. Available: \url{http://proceedings.mlr.press/v139/radford21a.html}
\BIBentrySTDinterwordspacing

\bibitem{diaz2018don}
N.~D{\'\i}az-Rodr{\'\i}guez, V.~Lomonaco, D.~Filliat, and D.~Maltoni, ``Don't forget, there is more than forgetting: new metrics for continual learning,'' \emph{arXiv preprint arXiv:1810.13166}, 2018.

\bibitem{hu2021lora}
E.~J. Hu, Y.~Shen, P.~Wallis, Z.~Allen-Zhu, Y.~Li, S.~Wang, L.~Wang, and W.~Chen, ``Lora: Low-rank adaptation of large language models,'' \emph{arXiv preprint arXiv:2106.09685}, 2021.

\end{thebibliography}


\begin{thebibliography}{1}
\bibitem{IEEEhowto:kopka}
H.~Kopka and P.~W. Daly, \emph{A Guide to \LaTeX}, 3rd~ed.\hskip 1em plus
  0.5em minus 0.4em\relax Harlow, England: Addison-Wesley, 1999.
\end{thebibliography}

\begin{IEEEbiography}[{\includegraphics[width=1.0in,height=1.26in,clip,keepaspectratio]{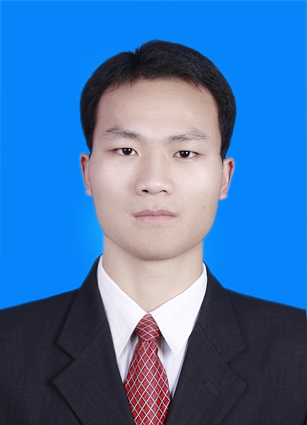}}]{Gan Sun} (S’19-M’20) is an associate professor in State Key Laboratory of Robotics, Shenyang Institute of Automation, Chinese Academy of Sciences. He received the B.S. degree from Shandong Agricultural University in 2013, the Ph.D. degree from Shenyang Institute of Automation, Chinese Academy of Sciences in 2020, and has been visiting Northeastern University from April 2018 to May 2019, Massachusetts Institute of Technology from June 2019 to November 2019. He also has some top-tier conference papers accepted at CVPR, ICCV, ECCV, AAAI, IJCAI, ICDM et al, and some top-tier journal papers accepted at TPAMI, TNNLS, TIP, TMM, TCSVT, Pattern Recognition et al. His current research interests include lifelong machine learning, multitask learning, medical data analysis, domain adaptation, deep learning and 3D computer vision.
\end{IEEEbiography}

%\vspace{-30pt}

\begin{IEEEbiography}[{\includegraphics[width=1in,height=1.25in,clip,keepaspectratio]{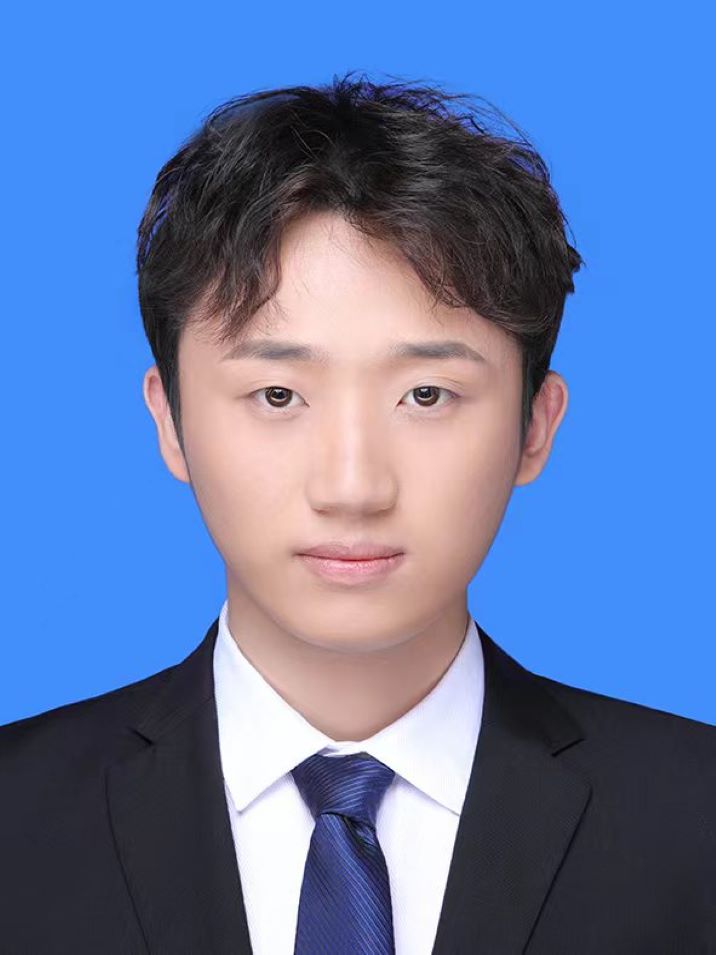}}]{Wenqi Liang} received the BS degree from Beijing Jiaotong University, Beijing, China, in 2022. He is currently working toward the MS degree in the State Key Laboratory of Robotics, Shenyang Institute of Automation, University of Chinese Academy of Sciences. He also has some top-tier conference papers accepted at IROS, ICCV et al. His current research interests include continual learning, generative AI and federated learning.
\end{IEEEbiography}

%\vspace{-30pt}

\begin{IEEEbiography}
[{\includegraphics[width=1in,height=1.25in, clip, keepaspectratio]{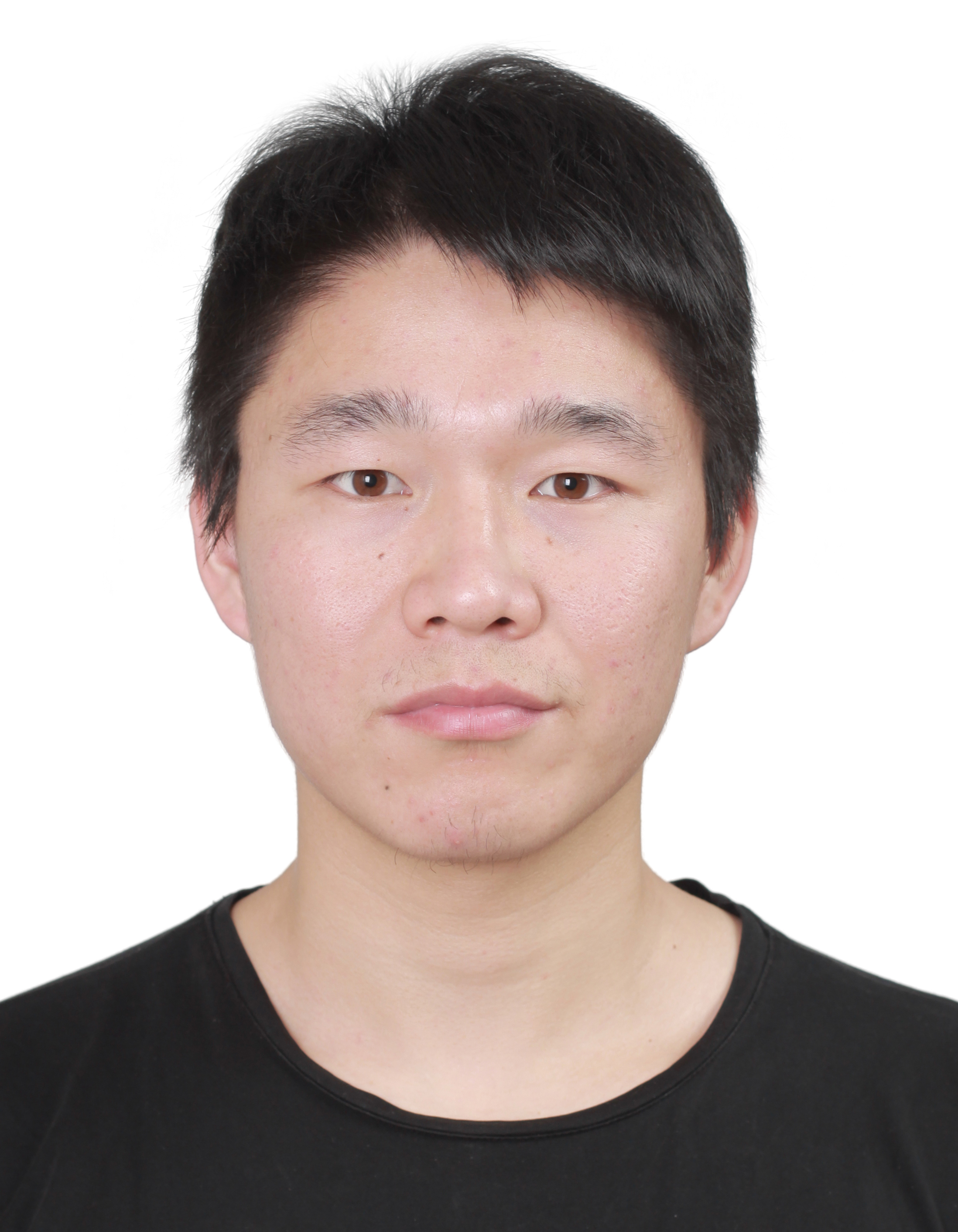}}]{Jiahua Dong} is currently a Ph. D candidate in State Key Laboratory of Robotics, Shenyang Institute of Automation, University of Chinese Academy of Sciences. He received the B.S. degree from Jilin University in 2017. He also has some top-tier conference papers accepted at CVPR, ICCV, ECCV, AAAI et al. His current research interests include transfer learning, robotic vision, medical image processing.
\end{IEEEbiography}

\begin{IEEEbiography}[{\includegraphics[width=1in,height=1.25in,clip,keepaspectratio]{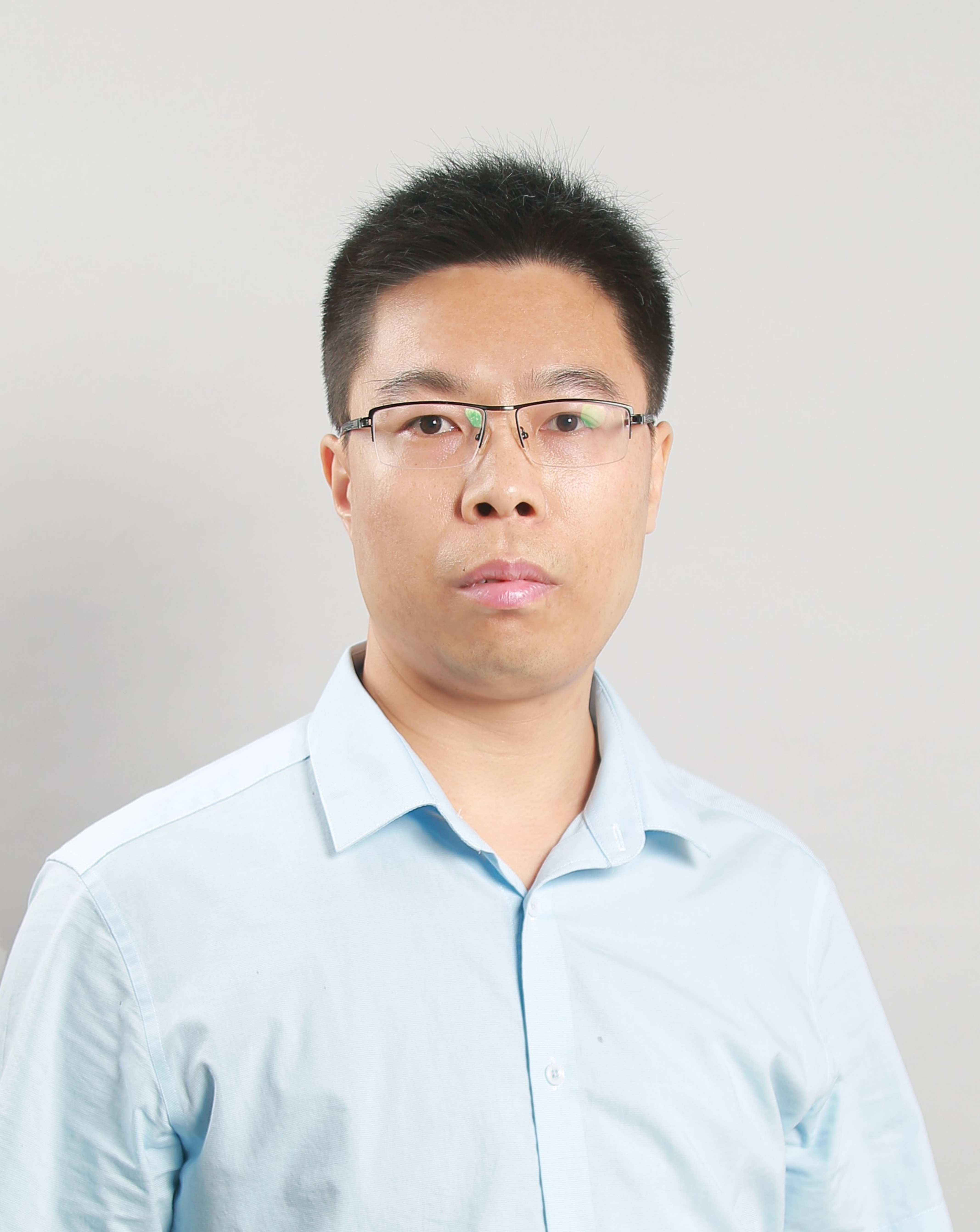}}]{Jun Li} (M'16) received the Ph.D. degree in pattern recognition and intelligence systems from the Nanjing University of Science and Technology in 2015. From Oct. 2012 to July 2013, he was a visiting student at Department of Statistics, Rutgers University, Piscataway, NJ, USA. From Dec. 2015 to Oct. 2018, he was a postdoctoral associate with the Department of Electrical and Computer Engineering, Northeastern University, Boston, MA, USA. From Nov. 2018 to Oct. 2019, he was a postdoctoral associate with the Institute of Medical Engineering and Science, Massachusetts Institute of Technology, Cambridge, MA, USA. He is currently a professor with the School of Computer Science and Engineering, Nanjing University of Science and Technology, Nanjing, China. He has served as a SPC/PC member for CVPR/ICCV/ECCV/ICML/NeurIPS/AAAI, and a reviewer for over 10 international journals such as IEEE TNNLS/TIP/TCYB/TCSVT. His research interests are computer vision and machine learning.
\end{IEEEbiography}

\begin{IEEEbiography}
[{\includegraphics[width=1in,height=1.25in, clip, keepaspectratio]{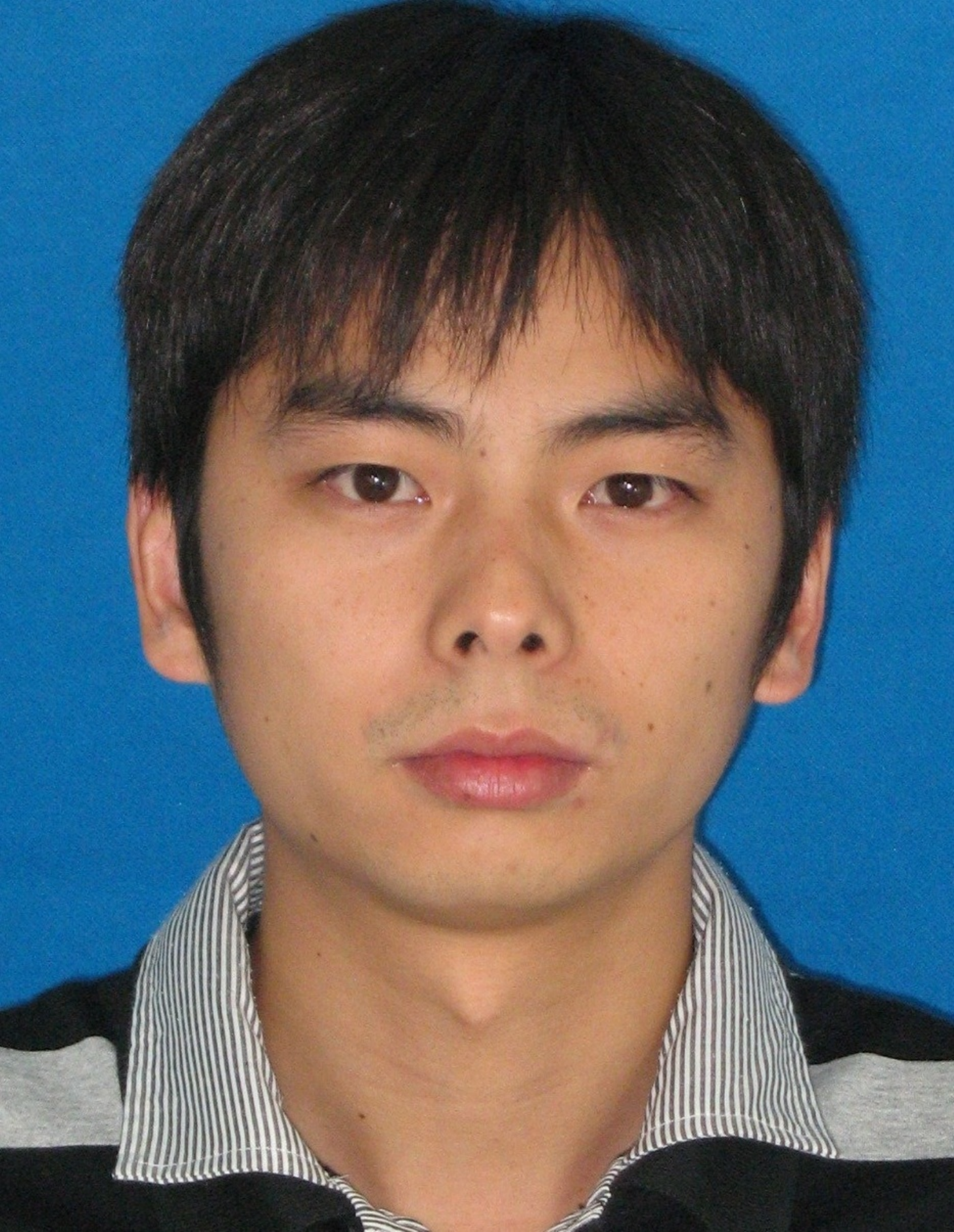}}]{Zhengming Ding}(S'14-M'18) received the B.Eng. degree in information security and the M.Eng. degree in computer software and theory from University of Electronic Science and Technology of China (UESTC), China, in 2010 and 2013, respectively. He received the Ph.D. degree from the Department of Electrical and Computer Engineering, Northeastern University, USA in 2018. He is a faculty member affiliated with Department of Computer Science, Tulane University since 2021. Prior that, he was a faculty member affiliated with Department of Computer, Information and Technology, Indiana University-Purdue University Indianapolis. His research interests include transfer learning, multi-view learning and deep learning. He received the National Institute of Justice Fellowship during 2016-2018. He was the recipients of the best paper award (SPIE 2016) and best paper candidate (ACM MM 2017). He is currently an Associate Editor of the Journal of Electronic Imaging (JEI) and IET Image Processing. He is a member of IEEE, ACM and AAAI.
\end{IEEEbiography}

\begin{IEEEbiography}[{\includegraphics[width=1in,height=1.25in,clip,keepaspectratio]{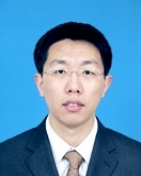}}]{Yang Cong} (S’09-M’11-SM’15) received the
B.Sc. degree from Northeast University in 2004 and the Ph.D. degree from the State Key Laboratory of Robotics, Chinese Academy of Sciences, in 2009.
From 2009 to 2011, he was a Research Fellow with the National University of Singapore (NUS) and Nanyang Technological University (NTU). He was
a Visiting Scholar with the University of Rochester. He was the professor until 2023 with Shenyang Institute of Automation, Chinese Academy of Sciences.
He is currently the full professor with South China University of Technology. He has authored over 80 technical articles. His current research interests include robot, computer vision, machine learning, multimedia, medical imaging and data mining. He has served on the editorial board of the several joural papers. He was a senior member of IEEE since 2015
\end{IEEEbiography}

\end{document}